\documentclass[10pt,twocolumn,letterpaper]{article}

\usepackage[pagenumbers]{cvpr} 

\usepackage{graphicx}
\usepackage{amsmath}
\usepackage{amssymb}
\usepackage{booktabs}
\usepackage{supertabular}
\usepackage{rotating}
\usepackage{caption}
\usepackage{subcaption}
\usepackage{graphicx}
\usepackage[pagebackref,breaklinks,colorlinks]{hyperref}

\usepackage[capitalize]{cleveref}
\crefname{section}{Sec.}{Secs.}
\Crefname{section}{Section}{Sections}
\Crefname{table}{Table}{Tables}
\crefname{table}{Tab.}{Tabs.}

\def\cvprPaperID{XXXX} 
\def\confName{CVPR}
\def\confYear{2022}

\begin{document}

\title{SPIN: Simplifying Polar Invariance for Neural networks \\ Application to vision-based irradiance forecasting}

%

{
\centering
\author{Quentin Paletta\textsuperscript{1,2}
\quad Anthony Hu\textsuperscript{1}
\quad Guillaume Arbod\textsuperscript{2}
\quad Philippe Blanc\textsuperscript{3}
\quad Joan Lasenby\textsuperscript{1}\\
\\
\textsuperscript{1}University of Cambridge, UK. \quad \textsuperscript{2}ENGIE Lab CRIGEN, France. \quad  \textsuperscript{3}MINES ParisTech, France.

}}


\maketitle

\begin{abstract}

Translational invariance induced by pooling operations is an inherent property of convolutional neural networks, which facilitates numerous computer vision tasks such as classification. Yet to leverage rotational invariant tasks, convolutional architectures require specific rotational invariant layers or extensive data augmentation to learn from diverse rotated versions of a given spatial configuration. Unwrapping the image into its polar coordinates provides a more explicit representation to train a convolutional architecture as the rotational invariance becomes translational, hence the visually distinct but otherwise equivalent rotated versions of a given scene can be learnt from a single image. We show with two common vision-based solar irradiance forecasting challenges (i.e. using ground-taken sky images or satellite images), that this preprocessing step significantly improves prediction results by standardising the scene representation, while decreasing training time by a factor of 4 compared to augmenting data with rotations. In addition, this transformation magnifies the area surrounding the centre of the rotation, leading to more accurate short-term irradiance predictions.

\end{abstract}

\section{Introduction}

The presence of clouds in the sky induces a short-term spatiotemporal variability in the production of solar energy. Their complex spatial distribution and temporal dynamics add to the difficulty of accurately predicting cloud displacements, and thus precisely forecasting solar energy yield, e.g. from photovoltaic or concentrated solar thermal electricity converting systems. Consequently, accounting for the high spatiotemporal solar irradiance variability caused by clouds is crucial for the integration of solar energy in the electric grid, or for off-grid hybrid systems, which use a fuel backup generator when the solar output is insufficient.
\vspace{0.5\baselineskip}

\begin{figure}
\centering
  \begin{minipage}[b]{0.155\textwidth}
    \includegraphics[width=0.92\textwidth]{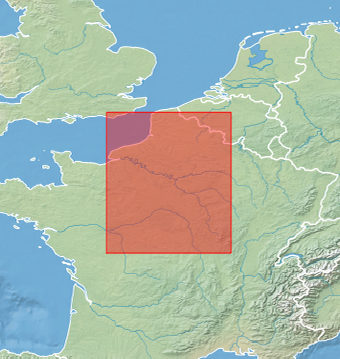}
  \end{minipage}
\begin{minipage}[b]{0.155\textwidth}
    \includegraphics[width=0.97\textwidth]{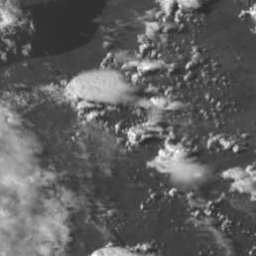}
    \label{fig:satellite_image}
  \end{minipage} 
  \begin{minipage}[b]{0.155\textwidth}
    \includegraphics[width=0.97\textwidth]{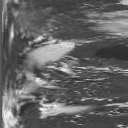}
  \end{minipage}
\vspace{-0.5\baselineskip}
\caption{Greyscale satellite image centred on SIRTA's Laboratory (48.713° N, 2.208° E)~\cite{eumetsatorganizationEUMETSATEuropeanOrganisation1991} and corresponding representation in polar coordinates.}
\label{fig:satellite_images}
\end{figure}

This short-term variability can be predicted from videos of the cloud cover taken by hemispherical sky cameras on the ground~\cite{peng3DCloudDetection2015a, blancShorttermForecastingHigh2017a} or geostationary satellites~\cite{hammerShorttermForecastingSolar1999}. The former provides local predictions with a high temporal resolution up to 20-min ahead, whereas the later gives hours-ahead predictions with a spatial resolution of up to 2-3 $\mathrm{km}^2$ (Figure~\ref{fig:satellite_images}). These vision-based approaches are better suited for prediction than time series from photovoltaic production monitoring or upstream pyranometric in-situ measurement (pyranometer: device used to measure the local solar irradiance), which have no intrinsic ability to predict sun covering by incoming clouds.

\begin{figure}
\centering
\begin{minipage}[b]{0.23\textwidth}
    \includegraphics[width=1.03\textwidth]{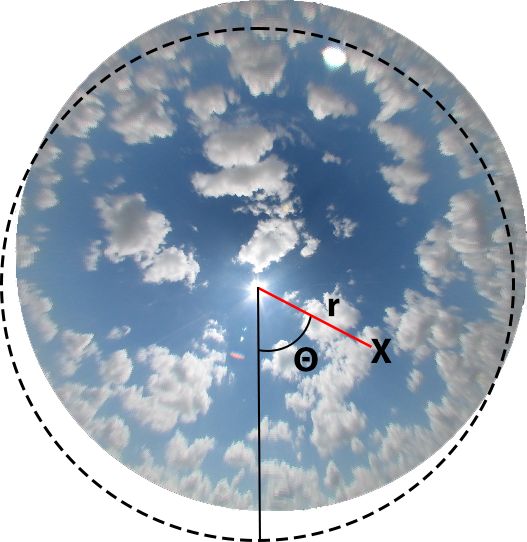}
    \label{fig:20180703112400_01}
  \end{minipage} 
  \;
  \begin{minipage}[b]{0.23\textwidth}
    \includegraphics[width=1.03\textwidth]{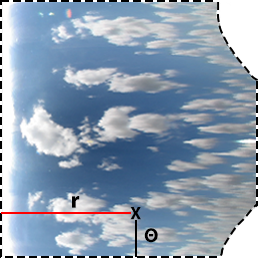}
    \label{fig:20180703112400_01_square}
  \end{minipage}
\vspace{-0.8\baselineskip}
\caption{Undistorted image of the sky taken by a fish-eye camera and its polar transformation (raw data are shared by SIRTA laboratory~\cite{haeffelinSIRTAGroundbasedAtmospheric2005}).}
\label{fig:stereo_polar}
\end{figure}

\vspace{0.5\baselineskip}

Invariances are key in computer vision as they can be integrated in the modelling as inductive biases to facilitate a learning process. For instance, in addition to being translationally equivariant due to weight sharing (a translated input results in similarly translated feature maps), a convolutional neural network using pooling layers is also translationally invariant, i.e. its prediction is unaffected by input translations. Disregarding the position of the object in the image significantly simplifies classification tasks, which share similar translational invariance properties, by avoiding learning to recognise an object for every possible position in the image.

\vspace{-0.8\baselineskip}
\paragraph{Definition} A task $\mathcal{T}$ is invariant with respect to an equivalence relation $\sim$~\footnote{A binary relation $\sim$ on a set $\mathcal{X}$ is said to be an equivalence relation, if and only if it is reflexive ($x \sim x$), symmetric ($x \sim x' \Leftrightarrow x' \sim x$) and transitive ($x \sim x'$ and $x' \sim x'' \Rightarrow x \sim x''$).} on $\mathcal{X}$ if:
\vspace{-0.6\baselineskip}
\begin{center}
$\forall x, x' \in \mathcal{X}$,  $x \sim x' \Rightarrow \mathcal{T}(x)=\mathcal{T}(x')$.
\end{center}

However, in some computer vision tasks such as solar energy forecasting, the position of objects is key. A principal component analysis on the learnt spatiotemporal representation of the sequence of past images has shown that the horizontal and vertical positions of the sun account for the 2nd and 4th principal components respectively~\cite{palettaECLIPSEEnvisioningCloud2021} (see Figure~\ref{fig:pca_components_examples_stereo}). The position of the sun in the sky and notably its zenith angle directly indicates the level of extra terrestrial irradiance and the amount of air mass crossed by the solar radiation. Hence, it is key to estimate the clear-sky downwelling irradiance (the component of radiation directed toward the earth's surface).

\vspace{0.5\baselineskip}
In addition, since the direct component accounts for most of the solar radiation, we expect the zenith angular distance from the centre of the sun to clouds to be of crucial importance to anticipate critical events. The azimuthal position of the clouds with respect the sun's position is, on the contrary, of much less importance. For these reasons, traditional techniques take the position of the sun in the image as a reference to read the spatial configuration and temporal dynamics of the cloud cover~\cite{marquezIntrahourDNIForecasting2013, quesada-ruizCloudtrackingMethodologyIntrahour2014a}. To some extent, the problem of irradiance forecasting from sky images can therefore be considered as rotationally invariant around the sun (polar invariance): the angle from which a cloud is approaching the sun has little impact on the resulting irradiance change. For similar reasons, local irradiance forecasting from satellite images can be approximately seen as invariant to rotation around a point of interest on the globe (a solar facility for instance).

\begin{figure*}
\centering
\begin{minipage}[b]{0.245\textwidth}
    \includegraphics[width=1\textwidth]{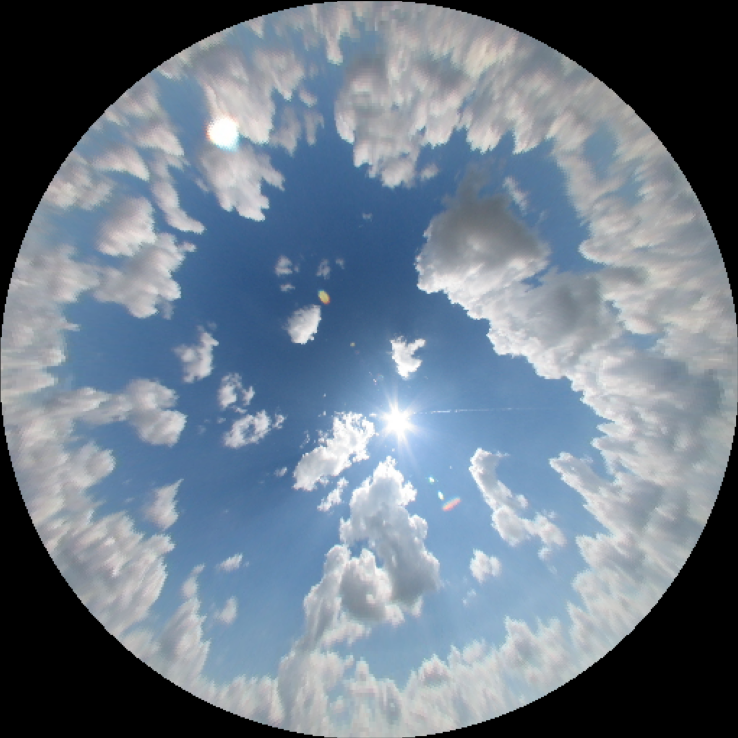}
    \label{fig:sky_image_undistorted}
  \end{minipage}
  \begin{minipage}[b]{0.245\textwidth}
    \includegraphics[width=1\textwidth]{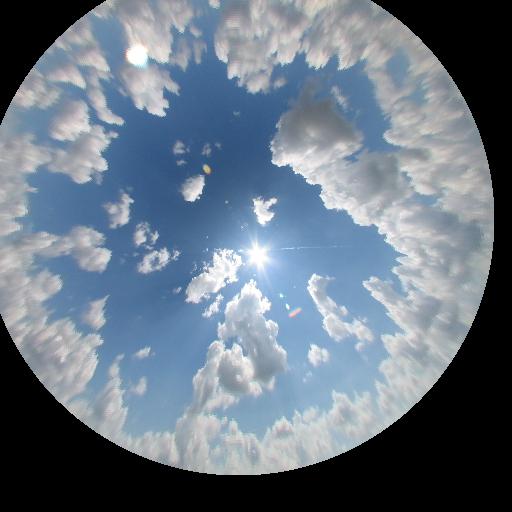}
    \label{fig:sky_image_target}
  \end{minipage}
   \begin{minipage}[b]{0.245\textwidth}
    \includegraphics[width=1\textwidth]{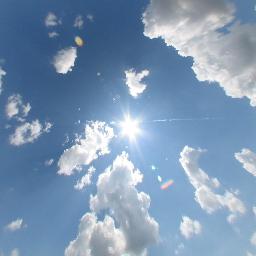}
    \label{fig:sky_image_nRBR}
  \end{minipage} 
  \begin{minipage}[b]{0.245\textwidth}
    \includegraphics[width=1\textwidth]{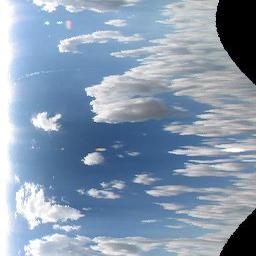}
    \label{fig:sky_image_segmente}
\end{minipage}

\begin{minipage}[b]{0.245\textwidth}
    \includegraphics[width=1\textwidth]{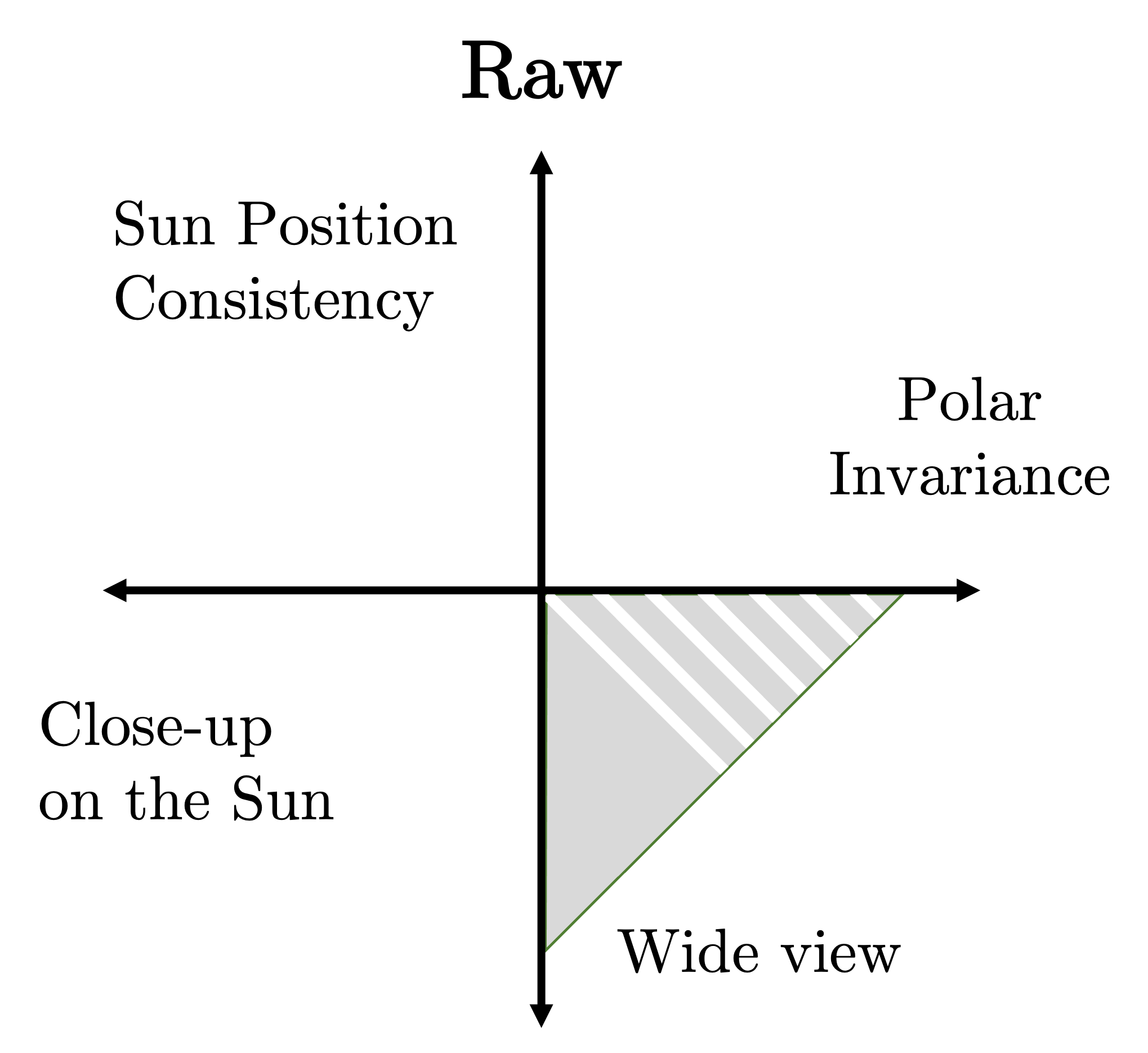}
    \label{fig:stereo_score_diagram}
  \end{minipage} 
\begin{minipage}[b]{0.245\textwidth}
    \includegraphics[width=1\textwidth]{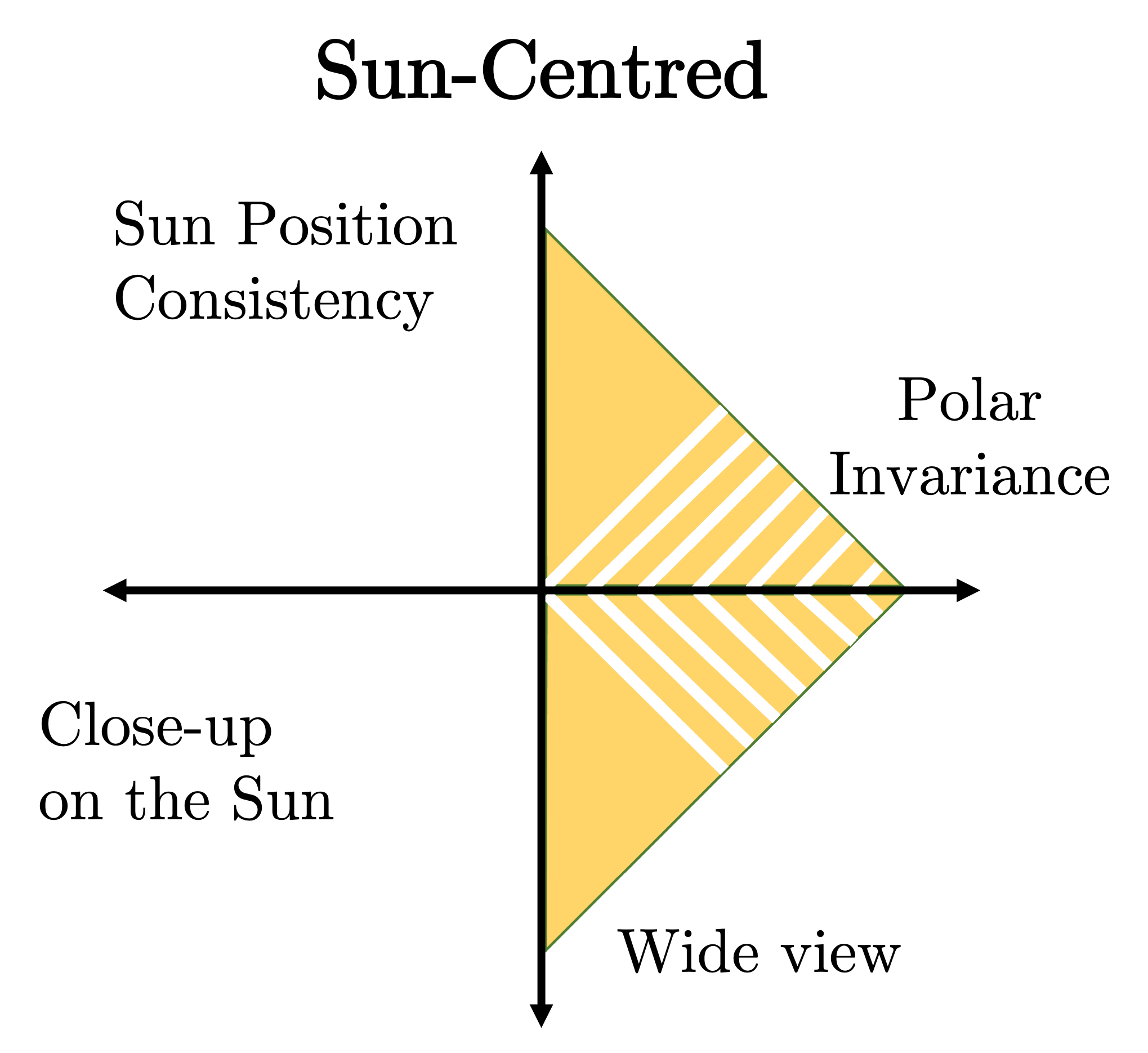}
    \label{fig:csa_score_diagram}
  \end{minipage}
  \begin{minipage}[b]{0.245\textwidth}
    \includegraphics[width=1\textwidth]{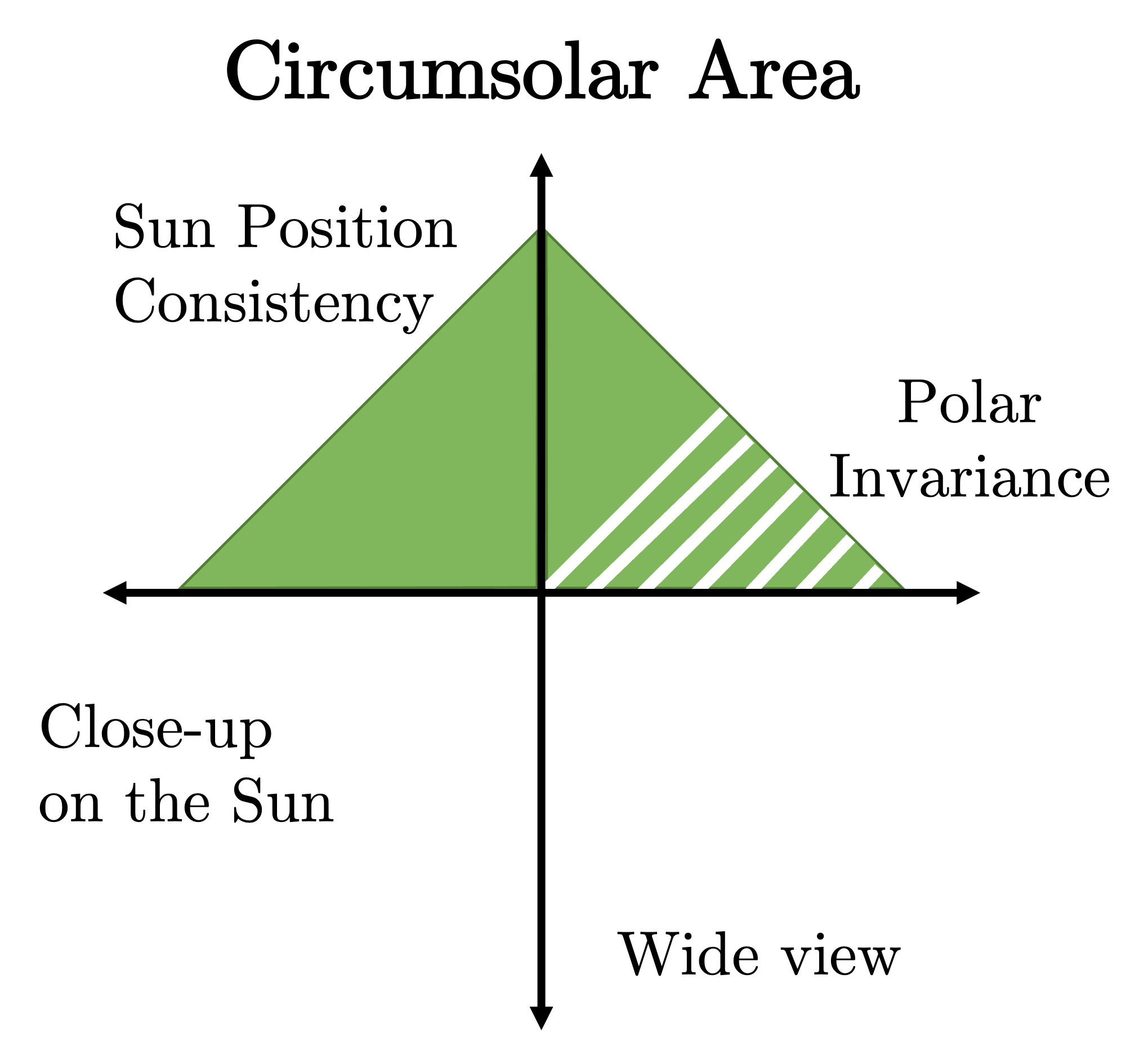}
    \label{fig:sun_centred_score_diagram}
  \end{minipage}
   \begin{minipage}[b]{0.245\textwidth}
    \includegraphics[width=1\textwidth]{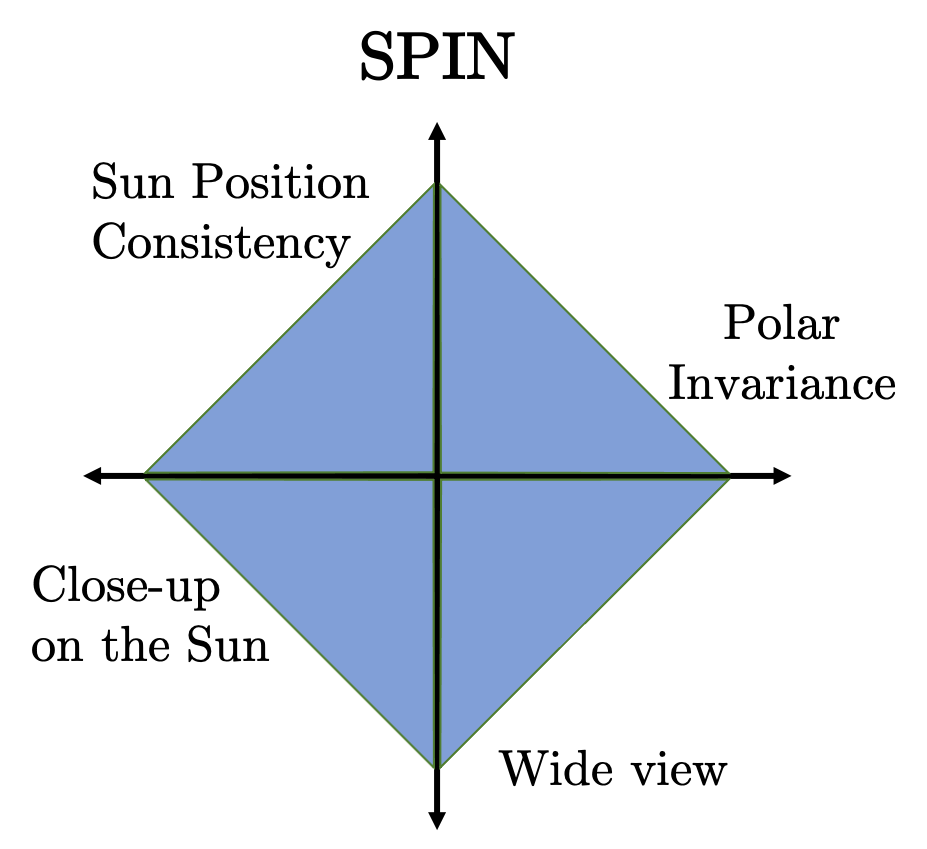}
    \label{fig:spin_score_diagram}
  \end{minipage}

\vspace{-1.2\baselineskip}
\caption{From left to right: 1. Raw undistorted image of the sky taken with a fish-eye camera, 2. Image centred on the sun's position, 3. Close up on the circumsolar area and 4. Polar coordinates. On top the different transformed images and below the corresponding properties diagrams (the white dashed area highlights the gain of leveraging the polar invariance with rotations) taking into account: 1.The polar invariance of the problem, 2. The central position of the sun in the scene, 3. The information close to the sun and 4. A wide view on the more distant part of the image.}
\label{fig:transformations}
\end{figure*}

\paragraph{Contributions} We introduce the SPIN method (Simplifying Polar Invariance for Neural Networks) to show the benefit for convolutional architectures of representing a rotationally invariant scene with polar coordinates in the context of video analysis. We propose the translation and vertical flip as data augmentation techniques for this polar representation but also temporal flip to leverage the temporal invariance of the problem. We compare it with other transformations and data augmentation approaches never before tried in deep irradiance forecasting from all-sky cameras: sun-centred images, a close-up on the circumsolar area and rotations to leverage the rotational invariance in the image by learning from more diverse cloud-sun spatial configurations. Two irradiance forecasting challenges using ground-taken sky images or satellite images are shown to greatly benefit from representing their corresponding scene with polar coordinates centred on the point of interest in the image, i.e. the sun in the sky image or a solar farm in the satellite image (see Figure~\ref{fig:satellite_images}).

\section{Prior work}

\paragraph{Vision-based solar irradiance forecasting} As well as traditional computer vision techniques to model cloud displacement based on cloud detection and tracking in a sequences of sky images~\cite{huangCloudMotionEstimation2013, pengHybridApproachEstimate2016, nouriNowcastingDNIMaps2018} or satellite observations~\cite{millerShorttermSolarIrradiance2018a}, recent studies have tried to learn the cloud cover dynamics from training end-to-end various types of neural networks from past sequences of hemispherical sky images~\cite{zhangDeepPhotovoltaicNowcasting2018, zhao3DCNNbasedFeatureExtraction2019, leguenDeepPhysicalModel2020a, kongHybridApproachesBased2020}. Benefiting from recent advances in deep learning (DL) and an increased availability of large sky images datasets~\cite{haeffelinSIRTAGroundbasedAtmospheric2005, kurtzVirtualSkyImager2017, pedroComprehensiveDatasetAccelerated2019, ntavelis2021skycam} and satellite data, past works have reported high forecasting scores~\cite{zhangDeepPhotovoltaicNowcasting2018, fengSolarNetSkyImagebased2020a, palettaBenchmarkingDeepLearning2021c}. The common setup is based on supervised learning: a convolutional architecture is trained to extract features from a sequence of past images (sky images or satellite images) and colocated/concomitant time series of pyranometric measurements, to predict future frames and/or future solar irradiance levels~\cite{palettaECLIPSEEnvisioningCloud2021, perezDeepLearningModel2021}. Based on the assumption that the neural network will be able to extract relevant features from raw images, input images are rarely preprocessed before being given to the model. In some studies, images taken by fish-eye cameras are unwrapped on a rectangular regular grid assuming for example, median cloud heights to limit the impact of the distortion on the scene representation. This provides cloud shape and trajectory consistency across the sequence of frames~\cite{palettaECLIPSEEnvisioningCloud2021}. Recently, some data augmentation techniques (e.g. noise injection, colour space transformations and mixing of images) have been applied to this forecasting task but no significant improvement was observed~\cite{nieResamplingDataAugmentation2021}.
\vspace{0.5\baselineskip}

\paragraph{Polar invariance} Although a rotational invariance around the point of interest is critical, it is not explicitly given to the neural network which has to comprehend a similar event (a cloud hiding the sun or covering the solar facility) multiple times for different positions of the cloud around the sun (or around the power plant in satellite images). A common but expensive way to learn this invariance would be to augment data by training the model on rotated versions of images around the sun (or the facility in the satellite image). Alternatively, one can turn this rotational invariance problem into a one of translational invariance by representing the scene with polar coordinates (see Figure~\ref{fig:stereo_polar})~\cite{berneckerContinuousShorttermIrradiance2014}. With this new configuration, the problem becomes invariant by translation along one axis (the argument coordinate), hence is more suited to analysis by convolutional layers. More than a data augmentation technique~\cite{salehinejadImageAugmentationUsing2018}, this transformation applied to rotationally invariant scenes is a more efficient way to learn a task invariance than data augmentation through rotations (which was not shown in common applications~\cite{jiangPolarCoordinateConvolutional2019a}) and does not require model tweaks~\cite{laptevTIPoolingTransformationInvariantPooling2016}. Additionally, this introduces a distortion in the scene representation resulting in an increased size of the region of interest (the circumsolar area in the sky image or the area surrounding the facility in the satellite image) relative to the rest of the image.

\section{Methodology}

\subsection{Datasets}

\paragraph{Irradiance data} Irradiance data were generated over a 3 year time window from 2017 to 2019 at SIRTA's lab~\cite{haeffelinSIRTAGroundbasedAtmospheric2005}. It was measured by a pyranometer and reported as a per minute average. Solar irradiance can be predicted from both sky images or satellite images with computer vision.

\paragraph{Sky Images} Sky images were generated over the same time window at SIRTA's lab. Images were taken by a hemispherical sky camera (EKO SRF-02) with a 2-min temporal resolution. To tackle the strong distortion induced by the fish-eye lens, images were undistorted~\cite{palettaTemporallyConsistentImagebased2020}. This provides cloud shape and trajectory consistency in consecutive frames as explained in the introduction. The segmentation of images into 5 classes (sky, clouds, sun, saturation and frame) following the method presented in \cite{palettaECLIPSEEnvisioningCloud2021} can be briefly summarised as follows: the image is first binary classified into cloud and sky pixels using the Hybrid Thresholding Algorithm (HYTA)~\cite{liHybridThresholdingAlgorithm2011, hasenbalgBenchmarkingSixCloud2020}, then the sun tracker~\cite{palettaTemporallyConsistentImagebased2020} is used to segment the sun if visible. Remaining saturated pixels are classified as saturation. Finally, RGB and segmented images are cropped and downscaled from $768 \times 1024$ to a $128 \times 128$ pixel resolution.

\vspace{-0.5\baselineskip}
\paragraph{Satellite Images} The set of greyscale satellite images used in this study was obtained from EUMETSAT~\cite{eumetsat1991} (Meteosat SEVIRI Rapid Scan image data~\footnote{https://navigator.eumetsat.int/product/EO:EUM:DAT:MSG:MSG15-RSS}). These $256 \times 256$ pixel images cover a $4.4^{\circ} \text{(latitude)}\times4.4^{\circ}$ (longitude) area centred on SIRTA's laboratory (48.713° N, 2.208° E) (Figure~\ref{fig:satellite_images}). The dataset comprises 148 days in 2017 (training set) and 198 days in 2018 (validation and test sets) with a 5-min temporal resolution.

\subsection{Image transformation and data augmentation}

\paragraph{Polar Transformation (SPIN)} The transformation of RGB and segmented images into their polar representations is depicted in Figure~\ref{fig:stereo_polar}. The centre of the rotation in raw images are respectively the centre of the image in satellite images and the centre of the sun in sky images, which is obtained from a tracking method~\cite{palettaTemporallyConsistentImagebased2020}. The radius $r$ ranges from $0$ to half of the original image's width $W$ in both cases and the polar angle $\theta$ from $0$ to $360^{\circ}$. In practice the original $W\times H \times 3$ image is first mapped to a $W/2 \times r \times 3$ image then downscaled to $128\times 128 \times 3$ (dimension of the input images in this study). Given the typical "U" shape trajectory of the sun in the sky image (see~\cite{palettaTemporallyConsistentImagebased2020}), the bottom half of the frame is more often visible in the polar transformation than the top half. To keep the resulting dark area in the corners of the resulting transformed image, the direction of the reference angle ($\theta=0$) is set to the downward vertical.

\vspace{0.5\baselineskip}
This transformation introduces several modifications in the representation of the scene, which might impact the performance of the learning (see Figure~\ref{fig:transformations}). First, the critical role of the sun position regardless of its visibility (e.g. hidden by a cloud) relative to the rest of the sky is taken into account to standardise the scene. Second, the rotational invariance of the problem around the sun is leveraged by turning it into a translational invariance problem. With this representation, the rotated versions of a given sun-clouds (or solar farm-clouds) spatial configuration can be learnt by a convolutional neural network (CNN) without the need to augment the data. Finally, the transformation distorts the representation of the scene leading to a magnifying effect on the circumsolar area (CSA; area directly surrounding the sun or the solar plant). To disentangle the impact of each aspect of the transformation, we compare the SPIN method with the following intermediate transformations:

\paragraph{Image centred on the Sun (Sky images)} To standardise the scene with respect to the position of the sun, we centre sky images on its position using the sun tracking algorithm (second panel in Figure~\ref{fig:transformations}).

\paragraph{Close-up on centre of the rotation} Building upon the sun-centred image transformation, we disentangle the magnifying effect on the CSA by training models on a close-up of the original image. The new representation depicted in the third panel of Figure~\ref{fig:transformations} corresponds to a fourth of the original image size.

\paragraph{Data augmentation} Finally, to leverage the rotational invariance of the problem, we perform random rotations of the input images to generate visually diverse samples from a single spatial configuration. This preprocessing step is applied to sun-centred images, the close-up on the CSA as well as on undistorted images to evaluate the effect of the resulting data augmentation. Although augmenting a polar coordinates image with vertical translations would generate redundant samples for a CNN, it provides some diversity by revealing the clouds split into two at the top and bottom edges of the image.

\vspace{0.5\baselineskip}
To compare all transformations keeping the model unchanged, each transformed image is eventually re-scaled to a $128\times 128 \times 3$ pixel resolution.

\subsection{Metrics}
\label{section:metrics}

{\bf Forecast Skill} Comparing the performance of forecasting methods evaluated in different conditions is problematic due to confounding effects of multiple variables: weather conditions, localisation, time of the year or of the day, etc. To partially overcome these limiting factors, a standard evaluation method in solar energy forecasting is to estimate the relative performance of a model with a reference mode~\cite{yangVerificationDeterministicSolar2020a}. The most common baseline is an adaptation of the \textit{persistence model} (PM) called the \textit{smart persistence model} (SPM). Whereas the PM predicts no irradiance changes over a given forecast window $\Delta T$ (Equation~\ref{equ:persistence}), the SPM takes into account diurnal changes of the extra-terrestrial irradiance given by a clear-sky model~\cite{blancHelioClimProjectSurface2011} (Equation~\ref{equ:smart_persistence}).
\vspace{0.5\baselineskip}

\vspace{-1.0\baselineskip}
\begin{equation}
   \hat{I}(t+\Delta T) = I(t)
   \label{equ:persistence}
\end{equation}
\vspace{-2.1\baselineskip}

\begin{equation}
   \hat{I}(t+\Delta T) = k_c(t) \, I_{clr}(t+\Delta T) \, \text{with} \, k_c(t) =  \frac{I(t)}{I_{clr}(t)}
   \label{equ:smart_persistence}
\end{equation}

\vspace{-0.2\baselineskip}

$I$ represents the measured irradiance and $\hat{I}$ is the predicted irradiance. $I_{clr}(t)$ is the corresponding clear-sky irradiance (without considering cloud) only accounting for the solar zenith angle and the optical transparency state of the atmosphere (or turbidiity). This is a modelled irradiance that is totally exogeneous from this study (McClear model~\cite{lefevreMcClearNewModel2013}). Given the errors of a model $\text{Error}_{model}$ and of the baseline $\text{Error}_{SPM}$ based on a given metric (RMSE, MSE, MAE, etc.), the \textit{forecast skill} (FS) is computed as follow:

\vspace{-0.7\baselineskip}
\begin{equation}
  \text{Forecast Skill} = 1-\frac{\text{Error}_{model}}{\text{Error}_{SPM}}
   \label{equ:FS}
\end{equation}

\vspace{0.12\baselineskip}
Although better able to generalise a model performance, the FS still has limitations~\cite{vallanceStandardizedProcedureAssess2017}. Particularly with DL approaches, architectures tend to face a frequent time lag correlated with low anticipation skills~\cite{palettaBenchmarkingDeepLearning2021c}. This greatly decreases their applicability to real world applications.
\vspace{0.5\baselineskip}

\begin{table*}[ht!]
\begin{center}
\begin{tabular}{lcccccccc}
\hline
 & & \multicolumn{3}{c}{RMSE $\downarrow$ [W/$\text{m}^2$] (Forecast Skill $\uparrow$ [\%])} & & \multicolumn{3}{c}{95\% Quantile $\downarrow$ [W/$\text{m}^2$]}\\

 Forecast Horizon & $\mid$ & 2-min & 6-min & 10-min & $\mid$ & 2-min & 6-min & 10-min \\
\hline\hline
\noalign{\vskip 2mm}
Smart Pers. & & 93.3 (0\%) & 129.7 (0\%) & 146.2 (0\%) & & 201.1 & 304.8 & 356.9 \\
\noalign{\vskip 2mm}
Raw && 85.4 (8.4\%) & 98.3 (24.2\%) & 110.4 (24.5\%) && 184.9 & 218.2 & 244.8 \\
Raw + Rotations && 78.6 (15.8\%) & 93.6 (27.8\%) & 105.5 (27.8\%) && 168.6 & 205.9 & 238.6 \\
\noalign{\vskip 2mm}
CSA && 80.4 (13.8\%) & 97.0 (25.5\%) & 109.8 (25.1\%) && 169.9 & 217.4 & 244.1 \\
CSA + Rotations && 77.1 (17.3\%) & 93.5 (27.9\%) & 107.3 (26.6\%) && 161.2 & 205.9 & 240.0 \\
\noalign{\vskip 2mm}
Sun-Centred && 81.0 (13.2\%) & 94.1 (27.4\%) & 106.6 (27.1\%) && 174.4 & 206.8 & 235.1 \\
Sun-Centred + Rotations && 74.9 (19.8\%) & 89.2 (31.2\%) & 102.7 (29.8\%) && 157.8 & 196.1 & 227.4 \\
\noalign{\vskip 2mm}
SPIN && 74.3 (20.4\%) & 89.3 (31.1\%) & 102.9 (29.6\%) && 156.6 & 196.6 & 230.3 \\
SPIN + Translations && \textbf{71.8 (23.1\%)} & \textbf{87.2 (32.8\%)} & \textbf{101.0 (30.9\%)} && \textbf{149.1} & \textbf{192.2} & \textbf{224.0} \\
\noalign{\vskip 1mm}
\hline
\end{tabular}
\end{center}
\vspace{-1\baselineskip}
\caption{Performance comparison of the different image transformations based on ECLIPSE 2, 6 and 10-min ahead irradiance predictions from sky images.}
\label{tab:results_eclipse}
\end{table*}

{\bf Temporal Distortion Index} 

To quantify this temporal misalignment, \cite{frias-paredesIntroducingTemporalDistortion2016} introduces the \textit{temporal distortion index} (TDI). Based on dynamic time warping, this metric quantifies temporal distortion by locally warping a times series (e.g. predictions of the model) to resemble another (e.g. ground truth). The resulting TDI is defined as a percentage relative to the maximal distortion. It indicates if the forecasting is, on average, late or in advance, and to what extent.

\vspace{0.5\baselineskip}

{\bf 95\% Quantile} Anticipating sudden events corresponding to large irradiance shifts is critical to mitigate the effect of the solar production variability. In a hybrid power plant configuration, for instance solar panels with a diesel generator, the fossil fuel backup has to warm up for a few minutes before being able to cover a low solar yield and get synchronized to the local frequency of the isolated grid. To assess the performance of the model on these extreme events, we report the 95\% quantile of the sorted list of absolute errors.

\subsection{Models}

The different transformations presented in this study are evaluated based on the performance of three DL irradiance forecasting models: a CNN~\cite{palettaConvolutionalNeuralNetworks2020}, a convolutional long short-term memory network (ConvLSTM)~\cite{palettaBenchmarkingDeepLearning2021c} network and ECLIPSE~\cite{palettaECLIPSEEnvisioningCloud2021}. The CNN and ConvLSTM designs are similar: a spatio-temporal representation of the input sequence of past images is learnt through a set of convolutional layers, pooling layers and recurrent modules (for the ConvLSTM). It is then decoded into a single future irradiance value. Instead of predicting a single irradiance value, ECLIPSE is trained end-to-end to recursively predict the sequence of future segmented images and corresponding irradiance levels (Table~\ref{tab:models}). This state-of-the-art design was shown to reduce forecast time lag, while improving quantitative performances~\cite{palettaECLIPSEEnvisioningCloud2021}.

\subsection{Model Training}

Samples with low solar elevation (less than 10° above the horizon) are not used in this study. The remaining samples are split in distinct sets as follows.

\paragraph{Sky Images} The model is trained on data from 2017 and 2018 (180,000 samples) and evaluated on even days from 2019 (20,000). The remaining odd days from 2019 are kept for testing (20,000). In addition, the TDM metric is tested on 200 windows randomly sampled from test days. Each window comprises 100 consecutive samples and is 3h20m long.

\paragraph{Satellite Images} The model is trained on data from 140 days in 2017 (16,000 samples) and validated on even days from 2018 (5,000). The remaining odd days from 2018 are retained for testing (5,000). In addition, the TDM metric is tested on 100 windows randomly sampled from test days. Each window comprises 50 consecutive samples and is 4h10m long.

\section{Results}

\subsection{Quantitative Performances}

The impact of each transformation on model predictions is evaluated through the performance of three models (CNN, ConvLSTM and ECLIPSE) on three forecast horizons (2, 6 and 10-min ahead). The scores of each configuration (Transformation $\times$ model $\times$ forecast horizon) are reported as their average over two trainings with different random initialisations (Tables~\ref{tab:results_table} and~\ref{tab:satellite_results_eclipse_long}). The results of ECLIPSE presented in Table~\ref{tab:results_eclipse} are analysed in greater detail as it is the best performing model on most indicators, but all three models display similar trends regarding image transformation and data augmentation.

\begin{figure}
\centering    
\includegraphics[width=0.48\textwidth]{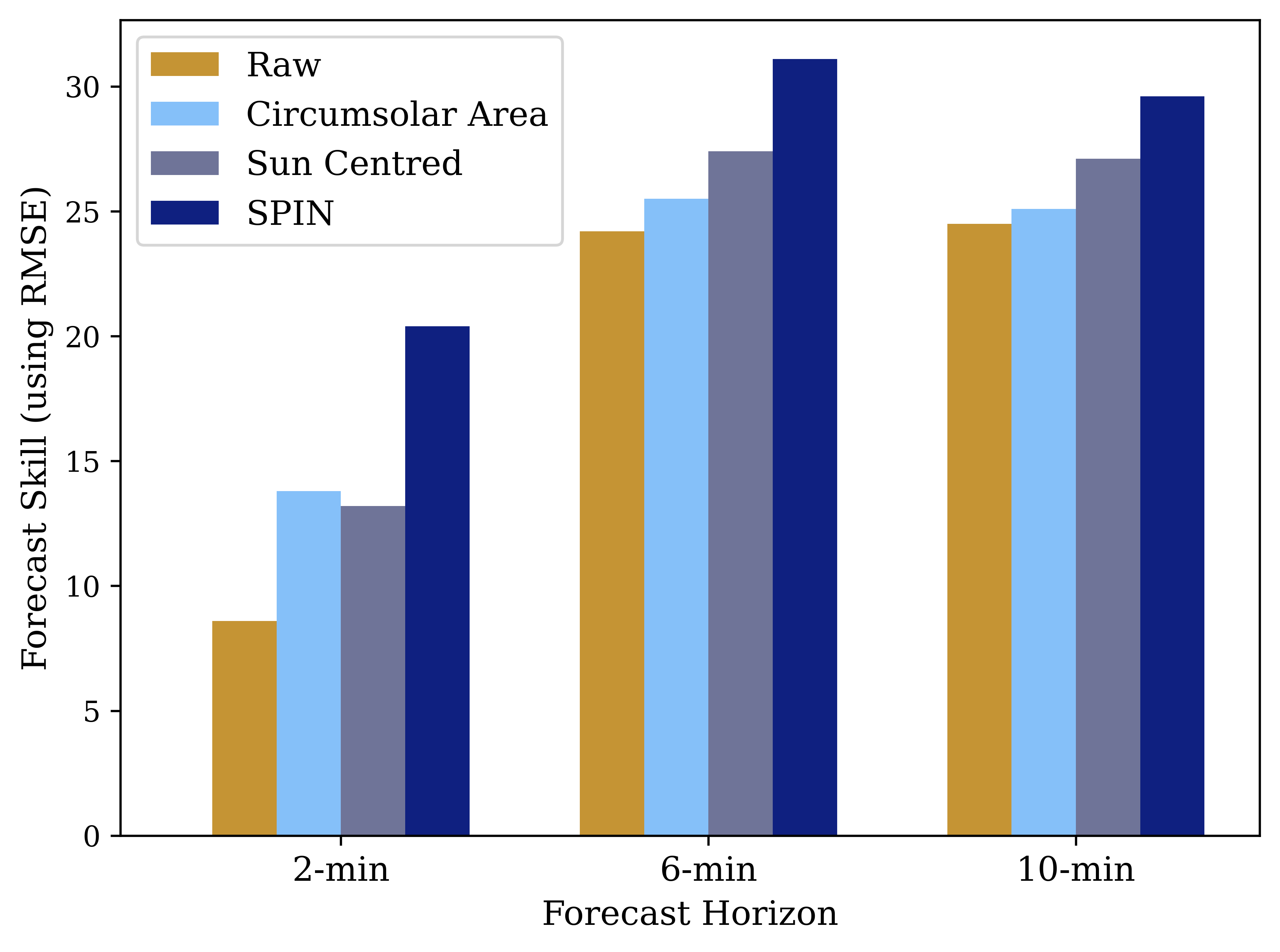}
\vspace{-1.1\baselineskip}
\caption{2, 6 and 10-min ahead FS of ECLIPSE predictions for different sky image transformations.}
\label{fig:bar_chart_eclipse_2}
\end{figure}

\vspace{0.5\baselineskip}
Although centring the image on the sun provides a considerable performance increase on all forecast horizons compared to the raw image (Figure~\ref{fig:bar_chart_eclipse_2}), the polar transformation provides the largest gain. Interestingly, the impact of each transformation on the model predictions depends on the horizon of the prediction. As the forecast horizon increases, the range of FS narrows: a difference of 12\% at 2-min (from 8.4 to 20.4\%), but only 5.1\% at 10-min (from 24.5 to 29.6\%). Ultimately, we expect all transformations to perform similarly on longer time horizons as current observations provide less information on the longer-term states of the cloud cover. This point is highlighted by the performance of ECLIPSE on the CSA transformation relative to the sun-centred image (Figure~\ref{fig:bar_chart_eclipse_2}). Whereas the CSA transformation leads to a better performance on the 2-min forecast (13.8 against 13.2\%), it is outperformed on longer time horizons by the sun-centred image (25.1 against 27.1\%), which provides more information on distant clouds. Similarly, the distortion observed in the SPIN scene representation seems to account for some of the FS gain on the 2-min ahead predictions: clouds surrounding the sun are more visible given the magnification of the CSA in polar coordinates.

\begin{figure}
\centering    
\includegraphics[width=0.48\textwidth]{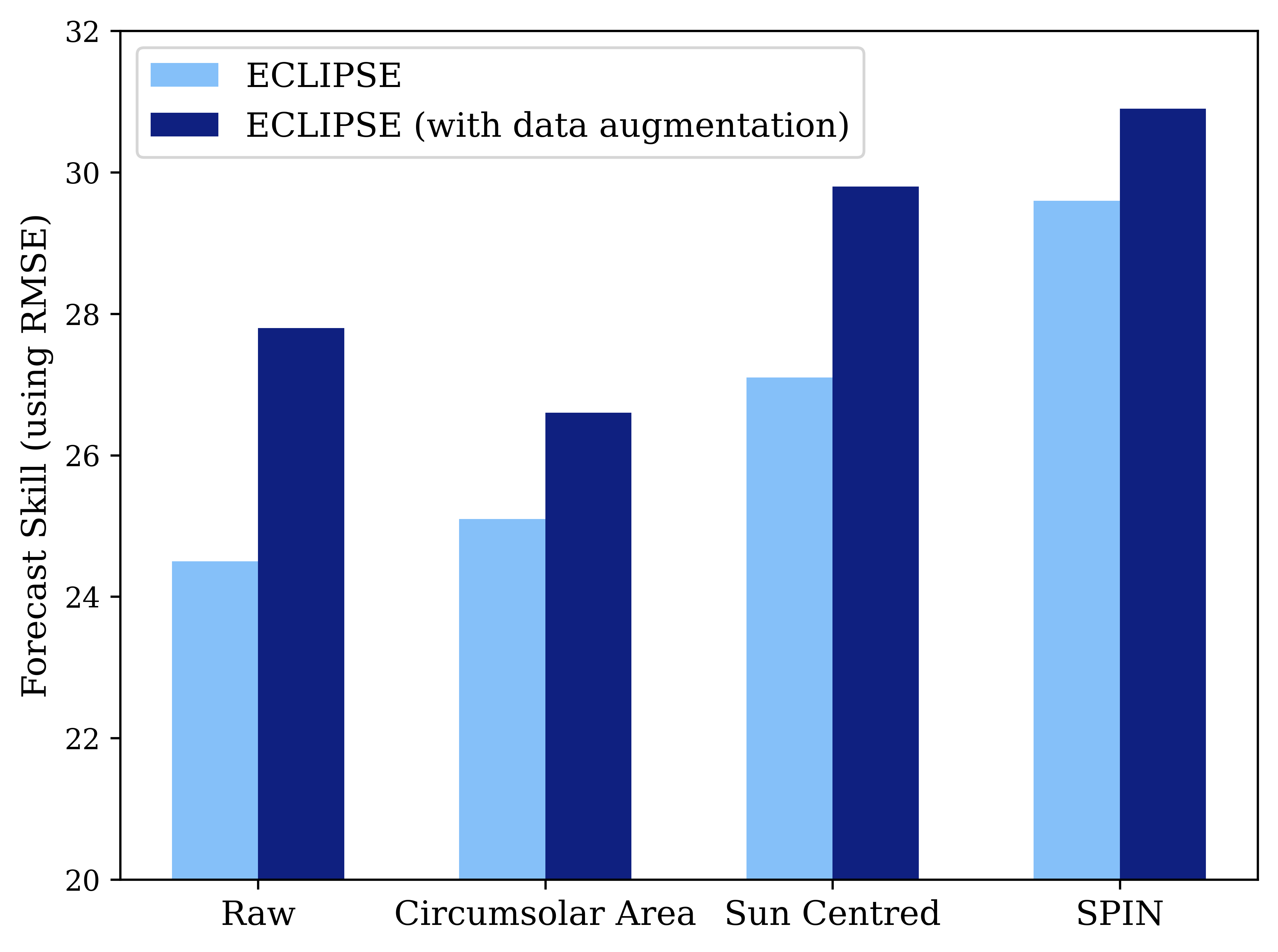}
\vspace{-1.1\baselineskip}
\caption{ECLIPSE 10-min ahead FS with and without data augmentation (rotation or translation) applied to the different transformations of sky images.}
\label{fig:bar_chart_eclipse_4}
\end{figure}

\begin{figure}
\centering    
\includegraphics[width=0.48\textwidth]{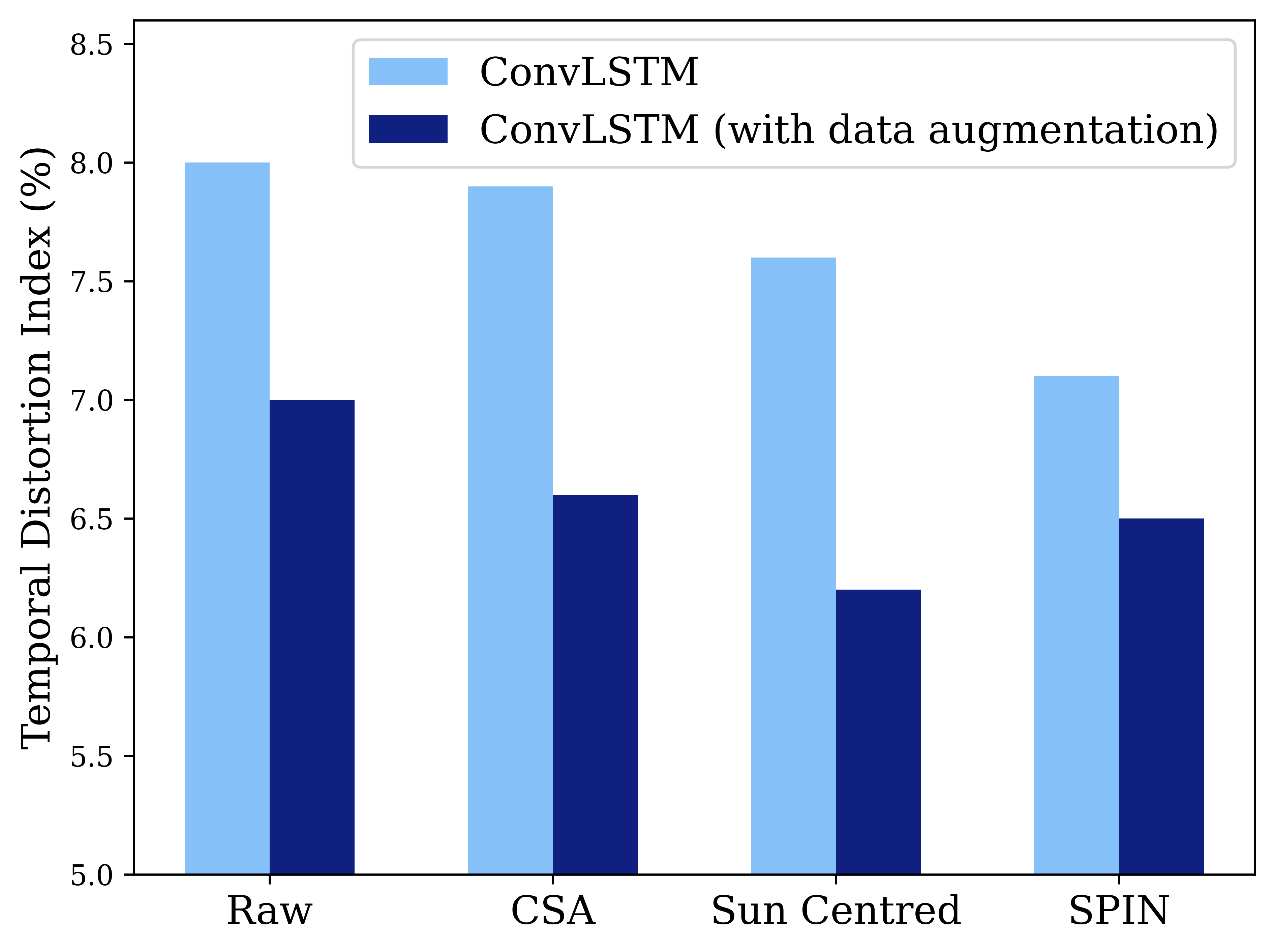}
\vspace{-1.1\baselineskip}
\caption{Time distortion index for the different sky image transformations averaged over the 2, 6 and 10-min ahead ConvLSTM's predictions.}
\label{fig:tdi_convlstm}
\end{figure}

\begin{figure}
\centering    
\includegraphics[width=0.48\textwidth]{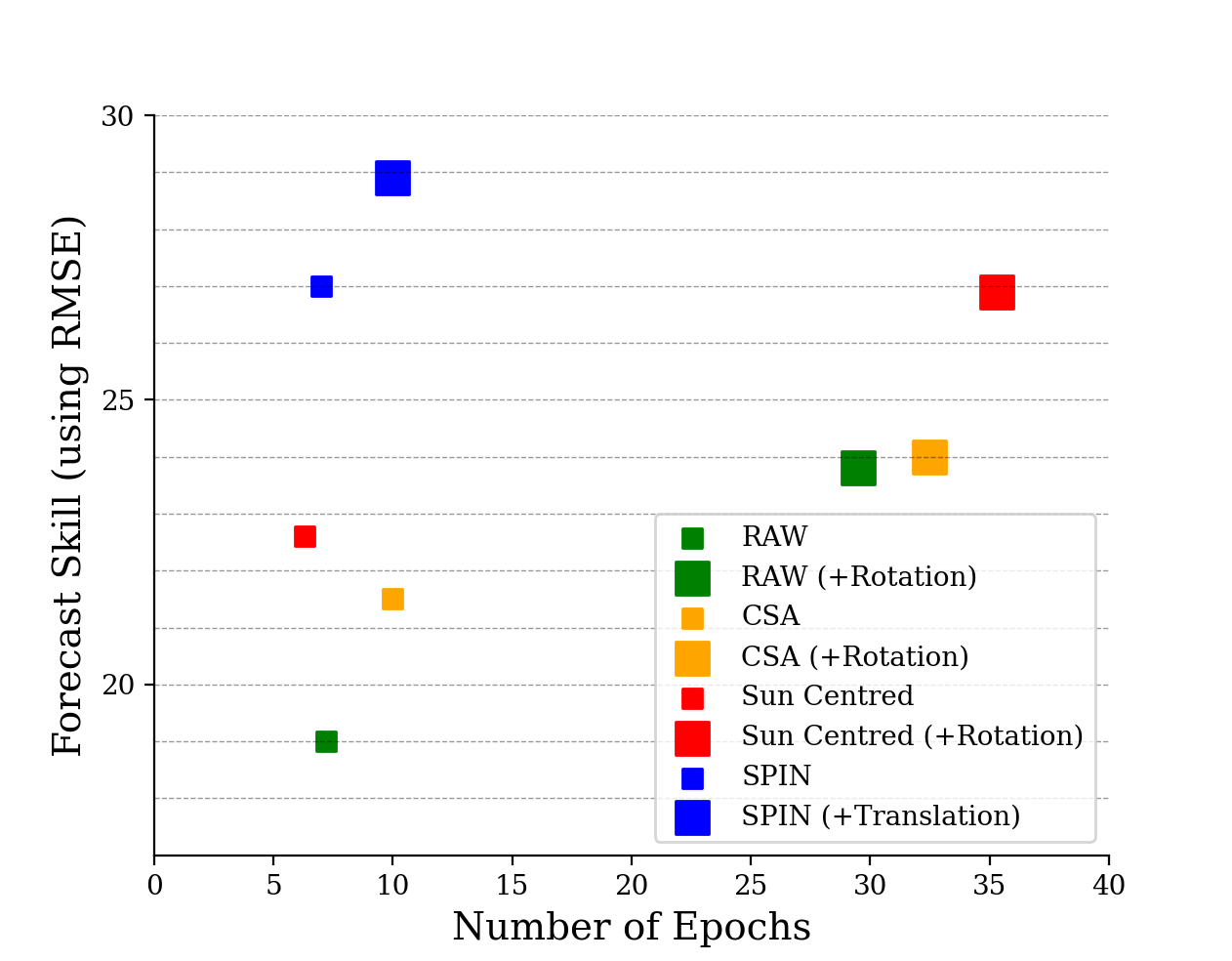}
\vspace{-0.8\baselineskip}
\caption{Impact of data augmentation on the training time for the different transformations averaged over the 2, 6 and 10-min ahead ECLIPSE predictions. Data augmentation has limited effect on SPIN's training time contrary to other transformations.}
\label{fig:eclipse_FS_epochs}
\end{figure}

\vspace{0.5\baselineskip}
Interestingly, data augmentation (rotation or vertical translation) consistently benefits predictions with an increased FS (Figure~\ref{fig:bar_chart_eclipse_4}) and a smaller temporal distortion (Figure~\ref{fig:tdi_convlstm}). 
Considering that the different rotated versions of a given sun-clouds configuration are learnt by a CNN from a single sample in polar coordinates, the benefit of translating the SPIN representation to reveal clouds split at the edge of the image is smaller than that of rotating the other transformations (Raw, Circumsolar and Sun-centred sky images) to learn from the resulting set of altered spatial configurations. This results in up to four times faster training for SPIN compared to the sun-centred representation with rotations for similar performances (Figure~\ref{fig:eclipse_FS_epochs}).

\subsection{Prediction Curves}

We compare the different transformations on the 2 and 10-min ahead prediction curves. Regarding the longer horizons, all transformations break the \textit{persistence barrier} (i.e. are able to foresee events before they happen and decrease time lag below the forecast horizon) except for the CSA (Figure~\ref{fig:prediction_curves_5lf}). This is understandable as a close-up on the sun provides little information on distant clouds which are more likely to impact longer-term temporal variability.

\vspace{0.5\baselineskip}
In addition, transformations magnifying the circumsolar area (CSA and SPIN) appear to induce a more accurate concomitant irradiance estimate, benefiting shorter term forecasting (2-min). In Figure~\ref{fig:prediction_curves_1lf} for instance, the CSA and SPIN predictions remain close to the ground truth, whereas the raw image and \textit{sun-centred} transformation lead to a high bias ($>100$ W/$\text{m}^2$) over large time windows.

\subsection{Visualisation of the Encoded Spatiotemporal Representation}

\begin{figure}[ht]
     \centering
\begin{subfigure}[b]{0.5\textwidth}
\centering
\begin{minipage}[b]{0.19\textwidth}
    \includegraphics[width=1\textwidth]{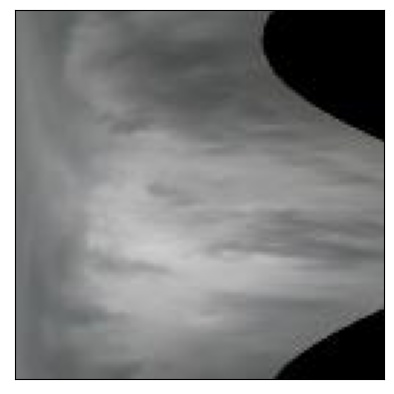}
  \end{minipage} 
\begin{minipage}[b]{0.19\textwidth}
    \includegraphics[width=1\textwidth]{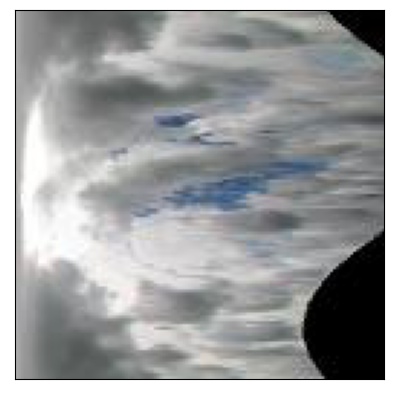}
  \end{minipage}
  \begin{minipage}[b]{0.19\textwidth}
    \includegraphics[width=1\textwidth]{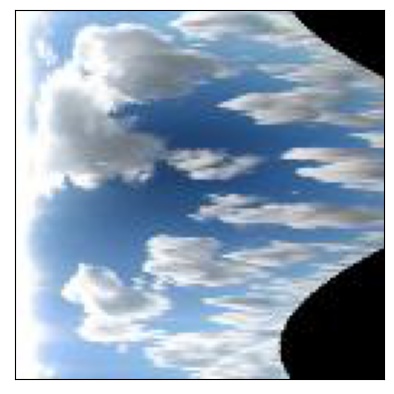}
  \end{minipage}
   \begin{minipage}[b]{0.19\textwidth}
    \includegraphics[width=1\textwidth]{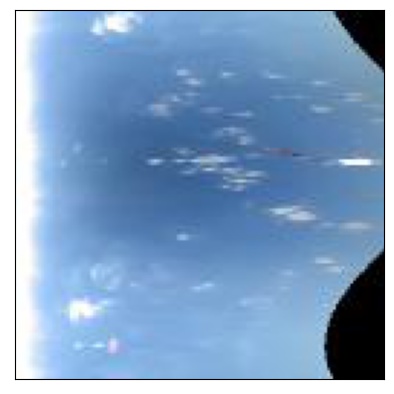}
  \end{minipage} 
  \begin{minipage}[b]{0.19\textwidth}
    \includegraphics[width=1\textwidth]{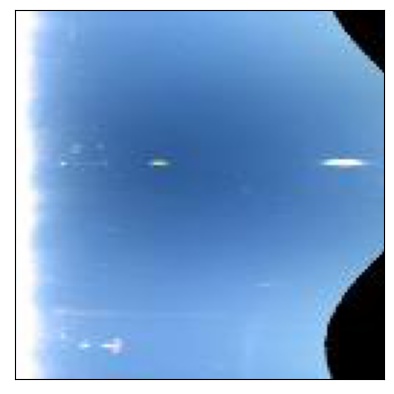}
\end{minipage}
\caption{PC1: Extent of the cloud coverage.}
\label{fig:pc1_polar}
\end{subfigure}
\hfill
\vspace{0.2\baselineskip}
\begin{subfigure}[b]{0.5\textwidth}
\centering
\begin{minipage}[b]{0.19\textwidth}
    \includegraphics[width=1\textwidth]{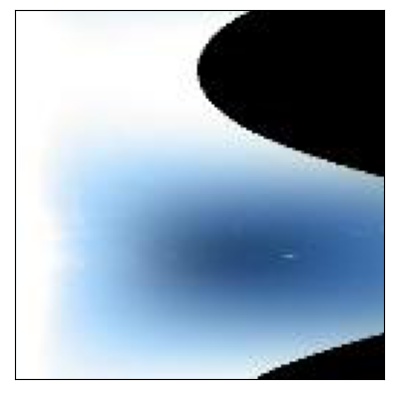}
  \end{minipage} 
\begin{minipage}[b]{0.19\textwidth}
    \includegraphics[width=1\textwidth]{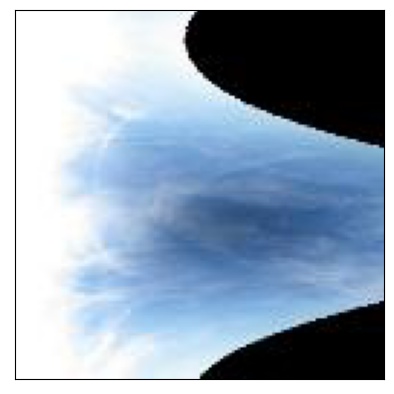}
  \end{minipage}
  \begin{minipage}[b]{0.19\textwidth}
    \includegraphics[width=1\textwidth]{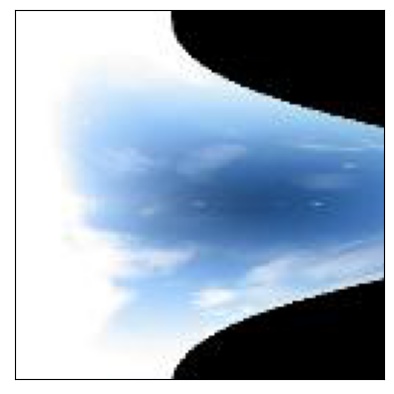}
  \end{minipage}
   \begin{minipage}[b]{0.19\textwidth}
    \includegraphics[width=1\textwidth]{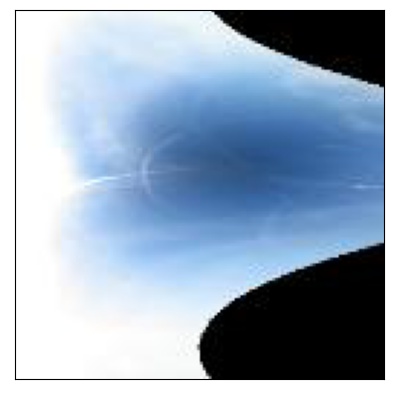}
  \end{minipage} 
  \begin{minipage}[b]{0.19\textwidth}
    \includegraphics[width=1\textwidth]{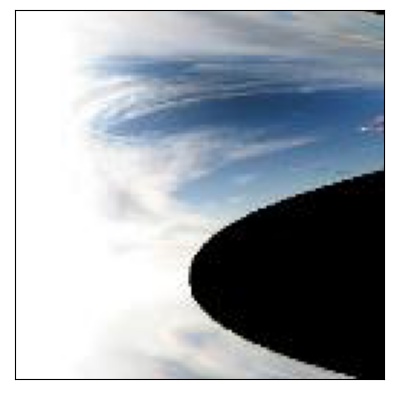}
\end{minipage}
\caption{PC2: Vertical position of the black area (frame) in the image. This indicates the position of the sun in the image, hence the level of clear-sky downwelling irradiance.}
\label{fig:pc2_polar}
\end{subfigure}
\hfill
\vspace{0.2\baselineskip}
\begin{subfigure}[b]{0.5\textwidth}
\centering
\begin{minipage}[b]{0.19\textwidth}
    \includegraphics[width=1\textwidth]{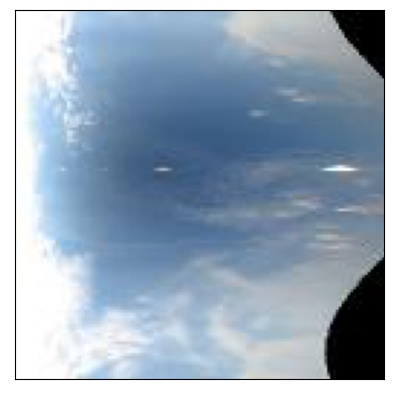}
  \end{minipage} 
\begin{minipage}[b]{0.19\textwidth}
    \includegraphics[width=1\textwidth]{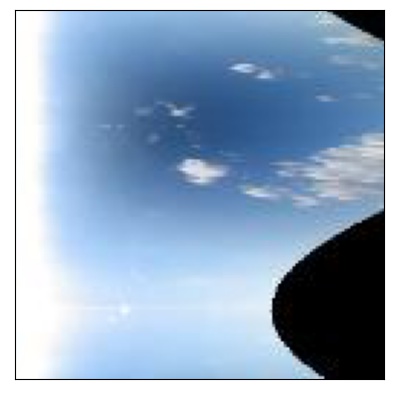}
  \end{minipage}
  \begin{minipage}[b]{0.19\textwidth}
    \includegraphics[width=1\textwidth]{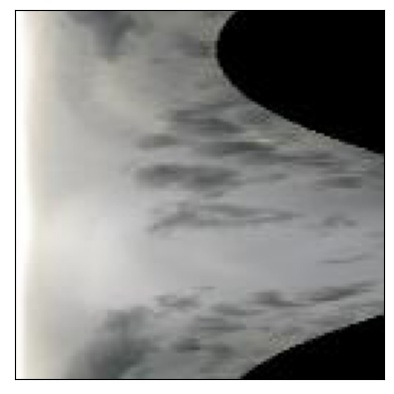}
  \end{minipage}
   \begin{minipage}[b]{0.19\textwidth}
    \includegraphics[width=1\textwidth]{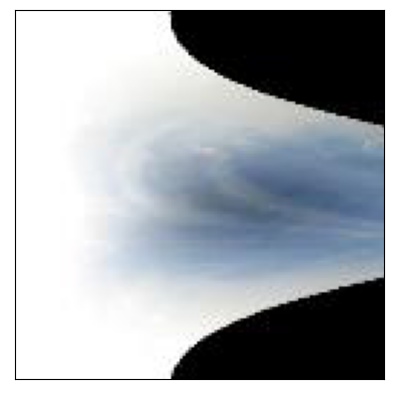}
  \end{minipage} 
  \begin{minipage}[b]{0.19\textwidth}
    \includegraphics[width=1\textwidth]{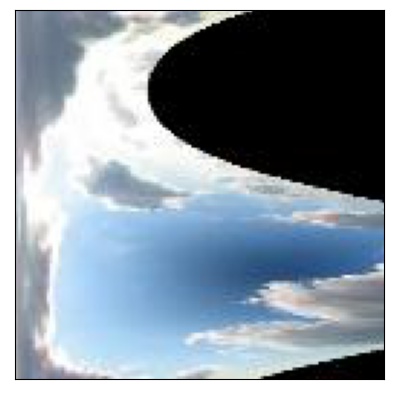}
\end{minipage}
\caption{PC3: Horizontal extent of the black area (frame) in the image. Similarly to PC2, this correlates with the sun position: the shorter the black tail, the further the sun from the horizon line, hence the higher the clear-sky downwelling irradiance. Contrary to PC2, this feature is invariant by vertical translation.}
\label{fig:pc3_polar}
\end{subfigure}
\hfill
\vspace{0.2\baselineskip}
\begin{subfigure}[b]{0.5\textwidth}
\centering
\begin{minipage}[b]{0.19\textwidth}
    \includegraphics[width=1\textwidth]{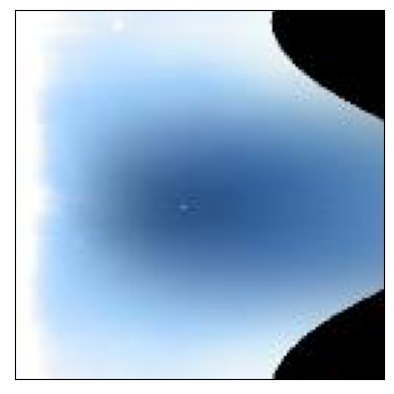}
  \end{minipage} 
\begin{minipage}[b]{0.19\textwidth}
    \includegraphics[width=1\textwidth]{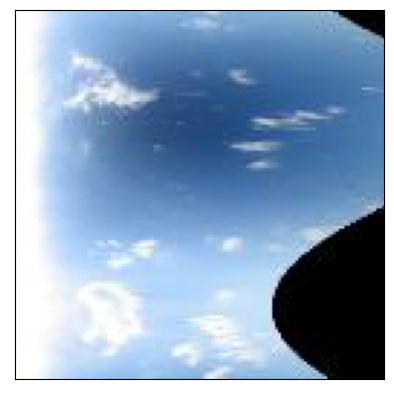}
  \end{minipage}
  \begin{minipage}[b]{0.19\textwidth}
    \includegraphics[width=1\textwidth]{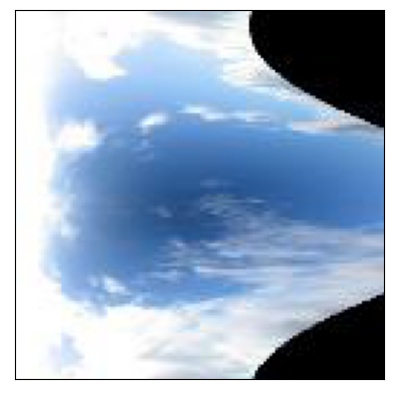}
  \end{minipage}
   \begin{minipage}[b]{0.19\textwidth}
    \includegraphics[width=1\textwidth]{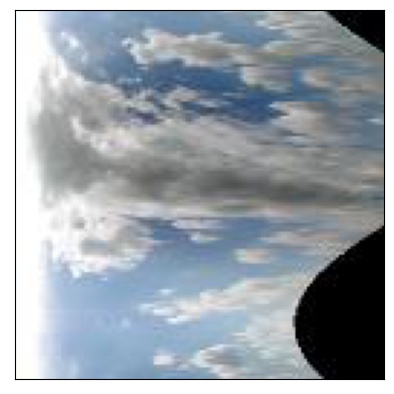}
  \end{minipage} 
  \begin{minipage}[b]{0.19\textwidth}
    \includegraphics[width=1\textwidth]{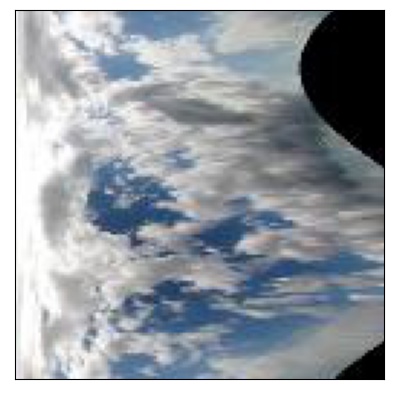}
\end{minipage}
\caption{PC4: Spatial variability of the cloud cover: from fully sunny (or fully cloudy) to partly cloudy.}
\label{fig:pc4_polar}
\end{subfigure}
\vspace{-0.1\baselineskip}
\caption{Four principal components of the spatio-temporal representation encoded by the model (with SPIN method). They respectively account for 8.6\%, 6.2\%, 4.4\% and 2.8\% of the variance. The variability of each component is illustrated with images draw from the distribution, from low to high values}
\label{fig:pca_components_examples_polar}
\end{figure}

We perform a principal component analysis (PCA) on the spatiotemporal representation of the past sequence of sky images (see Figure~\ref{fig:pca_components_examples_polar}). The first principal components are similar to those of raw sky images identified by~\cite{palettaECLIPSEEnvisioningCloud2021} (see Figure~\ref{fig:pca_components_examples_stereo}). In particular, the position of the sun seems to be extracted from the size and position of the black area (frame) in SPIN images. This indicates that the critical information of the sun position can be retrieved from polar representations, while it would be partly lost in rotated sky images.

\subsection{Satellite Images}

DL applied to satellite imagery is key for longer-term irradiance forecasting up to 1-2 hours, because of its rapid data processing at inference time compared to heavy numerical weather models, which need more time to reach steady-state simulation~\cite{bechDopplerRadarObservations2012}. In this study, ECLIPSE is trained on the 10 to 50-min ahead forecasts with a 10-min temporal resolution from greyscale satellite images, the corresponding cloud index map and irradiance measurement (included as a third input channel). Table~\ref{tab:ablation_study_aux_data} shows that models perform significantly better when based on the past irradiance levels, which can be done by inputting past measurements or by predicting the upcoming variation in irradiance levels instead of directly predicting absolute values.

\vspace{0.3\baselineskip}
Table~\ref{tab:satellite_results_eclipse_2} shows that both image transformation and data augmentation benefit irradiance forecasting in this new setting. The SPIN transformation leads to FS gains on predictions from both raw and close-up images on all horizons. Contrary to previous results on all-sky cameras, data augmentation on satellite images benefits all settings but the 10-min ahead predictions from polar coordinate representations (SPIN and SPIN close-up). This could be explained by the low level of cloud motion for shorter-term horizons, which could be learned without data augmentation.

\vspace{0.5\baselineskip}
\begin{table}[t]
\begin{center}
\begin{tabular}{lcccc}
\hline
 & & \multicolumn{3}{c}{Forecast Skill $\uparrow$ [\%]}\\
 Forecast Horizon & $\mid$ & 10-min & 30-min & 50-min\\
\hline\hline
\noalign{\vskip 1mm}
Raw && 8.5 & 6.5 & 6.3 \\
Raw (+R) && 12.1 & 12.2 & 13.7 \\
\noalign{\vskip 1mm}
SPIN && \textbf{12.2} & 12.5 & 14.1 \\
SPIN (+T) && 12.1 & \textbf{13.9} & \textbf{17.1} \\
\noalign{\vskip 1mm}
\hline
\hline
\noalign{\vskip 1mm}
Close-up && 10.3 & 10.6 & 11.8 \\
Close-up (+R) && 10.9 & 11.1 & \textbf{18.0} \\
\noalign{\vskip 1mm}
SPIN Close-up && \textbf{13.8} & 14.3 & 14.8 \\
SPIN Close-up (+T) && 13.6 & \textbf{14.4} & \textbf{18.0} \\
\noalign{\vskip 1mm}
\hline
\end{tabular}
\end{center}
\vspace{-1.1\baselineskip}
\caption{Performance benchmark of the different satellite image transformations based on ECLIPSE 10, 30 and 50-min ahead irradiance predictions averaged over a 5-min period (R : rotations, T : translations)}
\label{tab:satellite_results_eclipse_2}
\end{table}

\subsection{Temporal invariance}

\begin{figure}
\centering    
\includegraphics[width=0.48\textwidth]{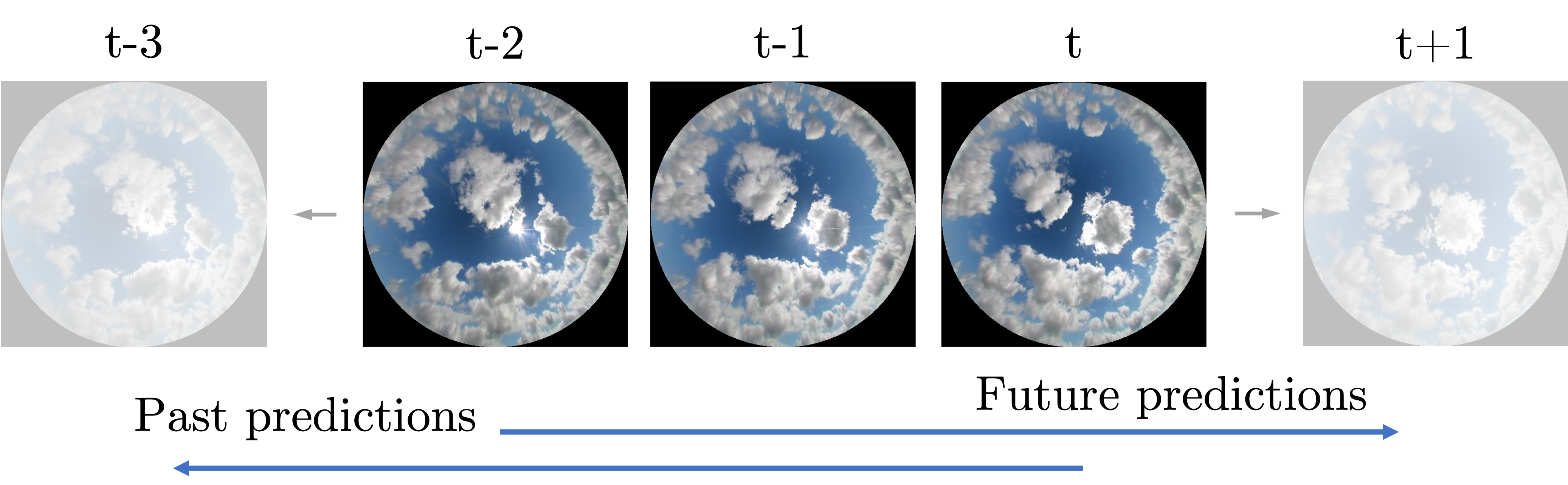}
\vspace{-1.3\baselineskip}
\caption{Temporal invariant task: predicting the past from a given sequence is as challenging as predicting the future. Training sequences are flipped with a probability of 0.5 to augment the data.}
\label{fig:temporal_invariance}
\end{figure}

Another key invariance in vision-based irradiance forecasting is the temporal invariance: from a given sequence of images, predicting past measurements is as challenging as predicting future measurements (Figure~\ref{fig:temporal_invariance}). To test that hypothesis, we augment the training data by reversing or not reversing input sequences with a probability of $50\%$ (both directions are seen by the model across multiple epochs).

\vspace{0.5\baselineskip}

\begin{table}[t]
\begin{center}
\begin{tabular}{lcccc}
\noalign{\vskip 2mm}
\hline
 & & \multicolumn{3}{c}{Forecast Skill $\uparrow$ [\%]}\\
 Forecast Horizon & $\mid$ & 2-min & 6-min & 10-min\\
\hline\hline
\noalign{\vskip 1mm}
SPIN && 21.4 & 32.2 & 30.2 \\
SPIN (+TF) && \textbf{23.2} & \textbf{32.9} & \textbf{30.6} \\

\noalign{\vskip 1mm}
SPIN (+T) && 22.2 & 32.5 & 30.5 \\
SPIN (+T\&TF) && \textbf{26.0} & \textbf{33.7} & \textbf{31.7} \\
\noalign{\vskip 1mm}
\hline
\end{tabular}
\end{center}
\vspace{-1\baselineskip}
\caption{Performance benchmark of the different sky image transformations based on ECLIPSE 2, 6 and 10-min ahead irradiance predictions averaged over a 5-min period (T : translations, TF : Temporal flipping). Due to missing data, samples with missing past instances were discarded, which is why the scores of the SPIN baselines are different from other experiments.}
\label{tab:sky_images_results_temporal_invariance}
\end{table}

\vspace{0.3\baselineskip}
Despite a 20 to 40\% increase in training time, results show a benefit on all forecast horizons of using temporal flipping as a data augmentation technique. Notably, the shorter the horizon the higher the FS gain: +1.8\% at 2-min, +0.7\% at 6-min and +0.4\% at 10-min (for SPIN with and without temporal invariance).

\subsection{Vertical flip}

We experimented vertical flipping as another type of data augmentation (training images were randomly flipped vertically with a probability of 50\%) combined with rotation and translation. The results presented in Table~\ref{tab:sky_images_results_flip} show that, although rotations and translations provide a more consistent gain, flipping images improves the baseline by about 1\% FS on all horizons. Combining vertical flipping with translations provides mixed results with no FS gain on the 6 and 10-min ahead horizons. Similarly to the temporal flip, the shorter the horizon, the higher the FS increase.

\begin{table}[t]
\begin{center}
\begin{tabular}{lcccc}
\hline
 & & \multicolumn{3}{c}{Forecast Skill $\uparrow$ [\%]}\\
 Forecast Horizon & $\mid$ & 2-min & 6-min & 10-min\\
\hline\hline
\noalign{\vskip 1mm}
SPIN && 20.4 & 31.1 & 29.6 \\
SPIN (+VF) && \textbf{23.8} & \textbf{32.3} & \textbf{30.4} \\
\noalign{\vskip 1mm}
SPIN (+T) && 23.1 & \textbf{32.8} & \textbf{30.9} \\
SPIN (+T\&VF) && \textbf{24.2} & 31.9 & 30.8 \\
\noalign{\vskip 1mm}
\hline
\end{tabular}
\end{center}
\vspace{-1\baselineskip}
\caption{Performance benchmark of the different sky image data augmentation techniques (T : translations, VF : Vertical flip) based on ECLIPSE 2, 6 and 10-min ahead irradiance predictions}
\label{tab:sky_images_results_flip}
\end{table}

\section{Discussion}

This study compares different image transformations in the context of rotationally invariant tasks. The characteristics of the different representations and the use of data augmentation leads to variable performance gains regarding forecast skill, clear-sky irradiance level estimation, temporal misalignment or training time. Regarding these aspects of irradiance forecasting, further investigations would focus on the impact of transformations' parameters such as the scaling factor for the CSA transform or the maximum radius for SPIN. In addition, we expect that combining the different representations as multiple input channels would further improve the predictions.

\vspace{0.5\baselineskip}
Furthermore, this learning strategy is likely to advance the generalisation properties of a model to observations from other locations. In a transfer learning context, cameras corresponding to a different dataset might not be oriented the same way, hence some spatial features such as the impact of the sun position on the irradiance level would be altered. However, the standardisation of the sun-clouds spatial configuration under the SPIN transformation could facilitate knowledge sharing.

\section{Conclusion}

By representing a scene with polar coordinates, the rotational invariance of a problem becomes translational. Thus, the visually diverse but equivalent spatial configurations obtained by rotating a given image are learnt by convolutional architectures from a single representation in polar coordinates. Applying this preprocessing step to two polar invariant computer vision-based irradiance forecasting problems, from sky cameras or satellite imagery, provided a significant performance increase compared to the single use of raw data without data transformation or data augmentation with rotations. Models learning from this representation train faster while decreasing temporal misalignment, a key aspect of forecasting tasks. In addition, the SPIN method is likely to facilitate knowledge sharing by standardising the representation of scenes observed with different camera setups.

\newpage

{\bf Acknowledgements} The authors acknowledge SIRTA and EUMETSAT for providing the data used in this study. We are grateful to Gisela Lechuga, Aleksandra Marconi and Marcos Gomes-Borges for their technical assistance and valuable comments on the manuscript. This research was supported by ENGIE Lab CRIGEN, EPSRC and the University of Cambridge.

{\small
\bibliographystyle{ieee_fullname}
\bibliography{MyLibrary}
}

\newpage

\appendix

\renewcommand\thefigure{\thesection.\arabic{figure}}
\renewcommand\thetable{\thesection.\arabic{table}}
\renewcommand\thetable{\thesection.\arabic{equation}}

\newpage
\onecolumn

\section{Dataset Balance - Sky Images}
\label{section:dataset_balance}
\setcounter{figure}{0}
\setcounter{table}{0}

\begin{figure}[h!]
\centering
\begin{minipage}[b]{0.33\textwidth}
    \includegraphics[width=1\textwidth]{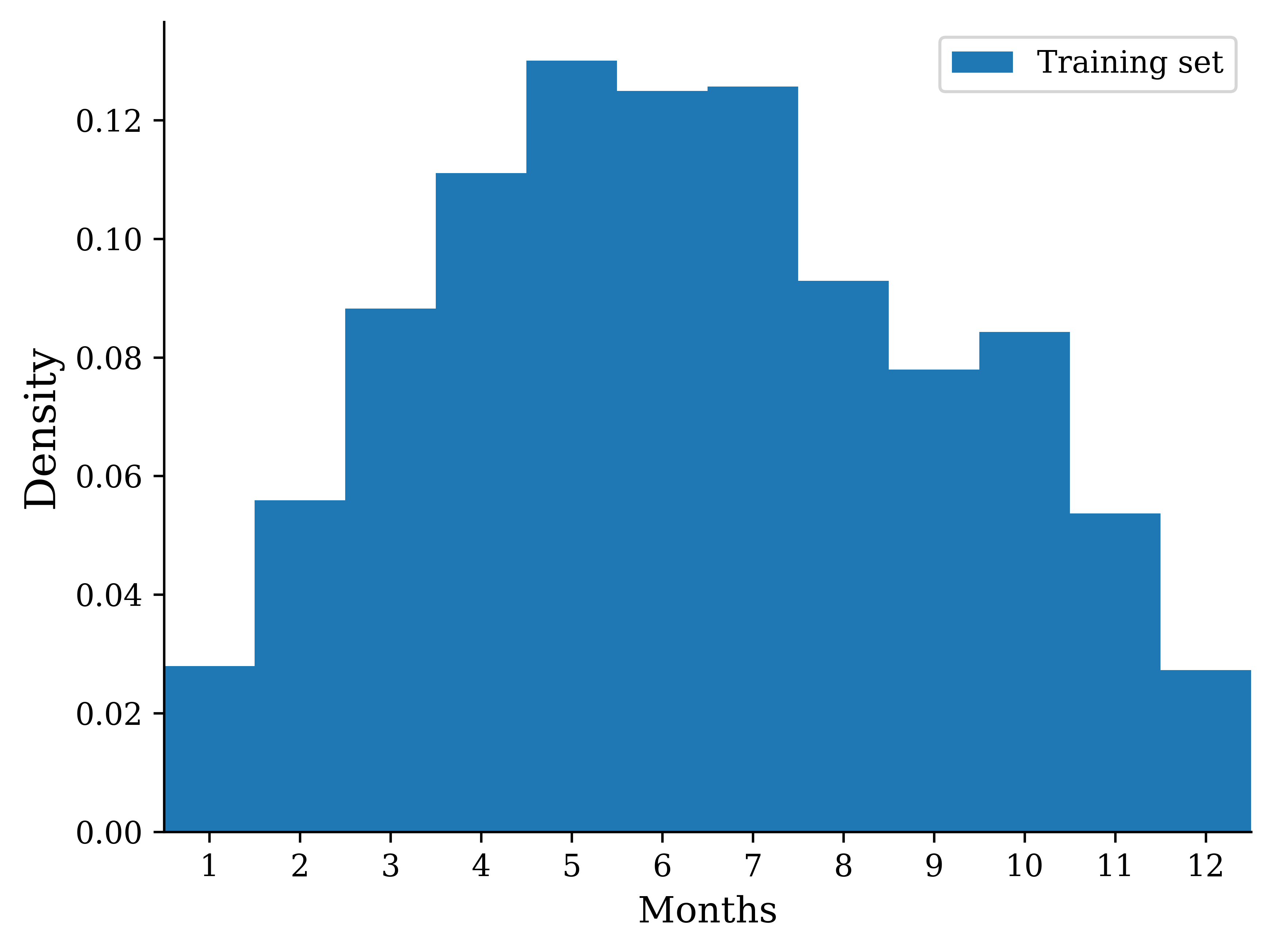}
    \label{fig:hist_months_train}
  \end{minipage} 
  \begin{minipage}[b]{0.33\textwidth}
    \includegraphics[width=1\textwidth]{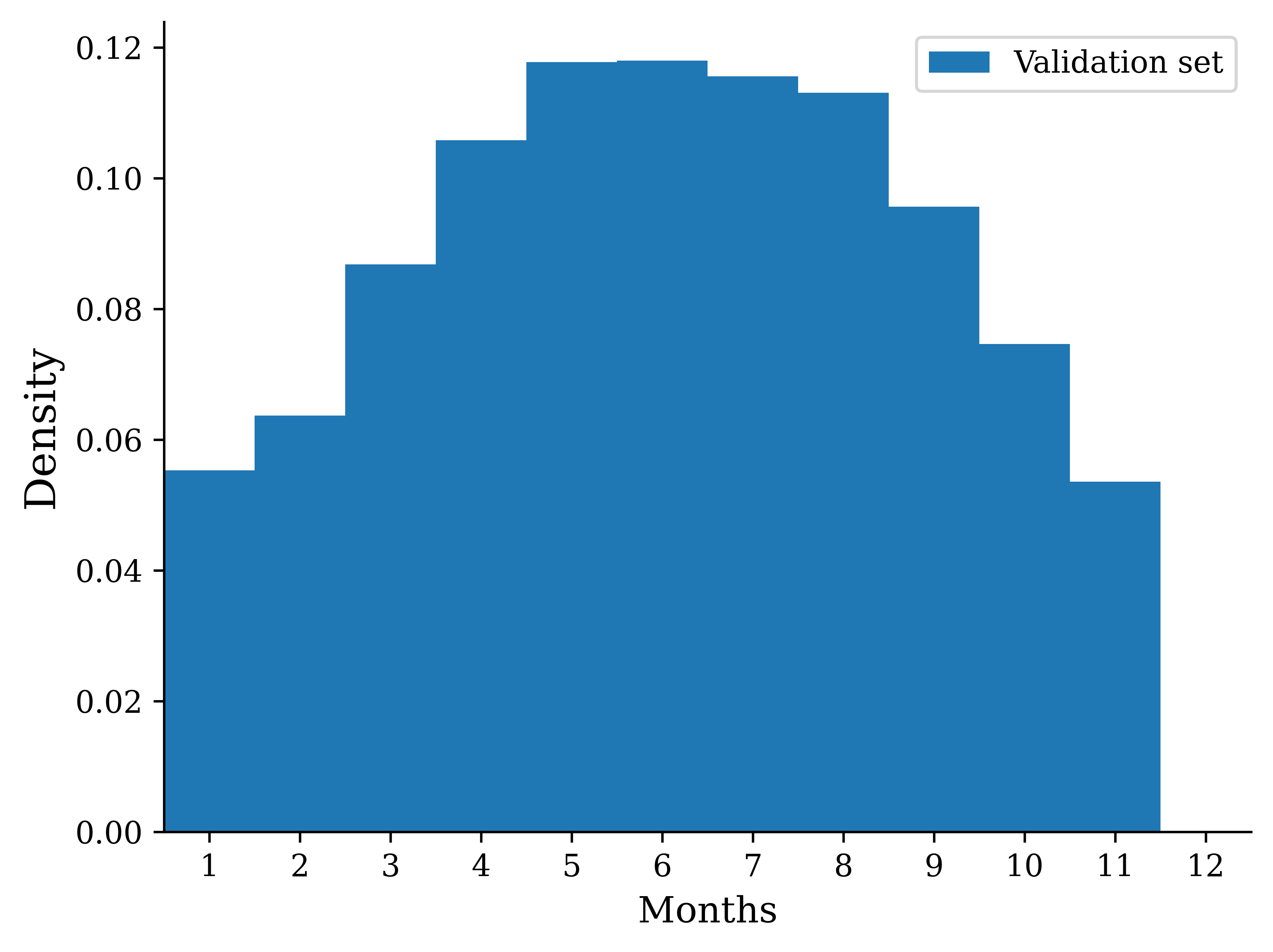}
    \label{fig:hist_months_val}
  \end{minipage}
  \begin{minipage}[b]{0.33\textwidth}
    \includegraphics[width=1\textwidth]{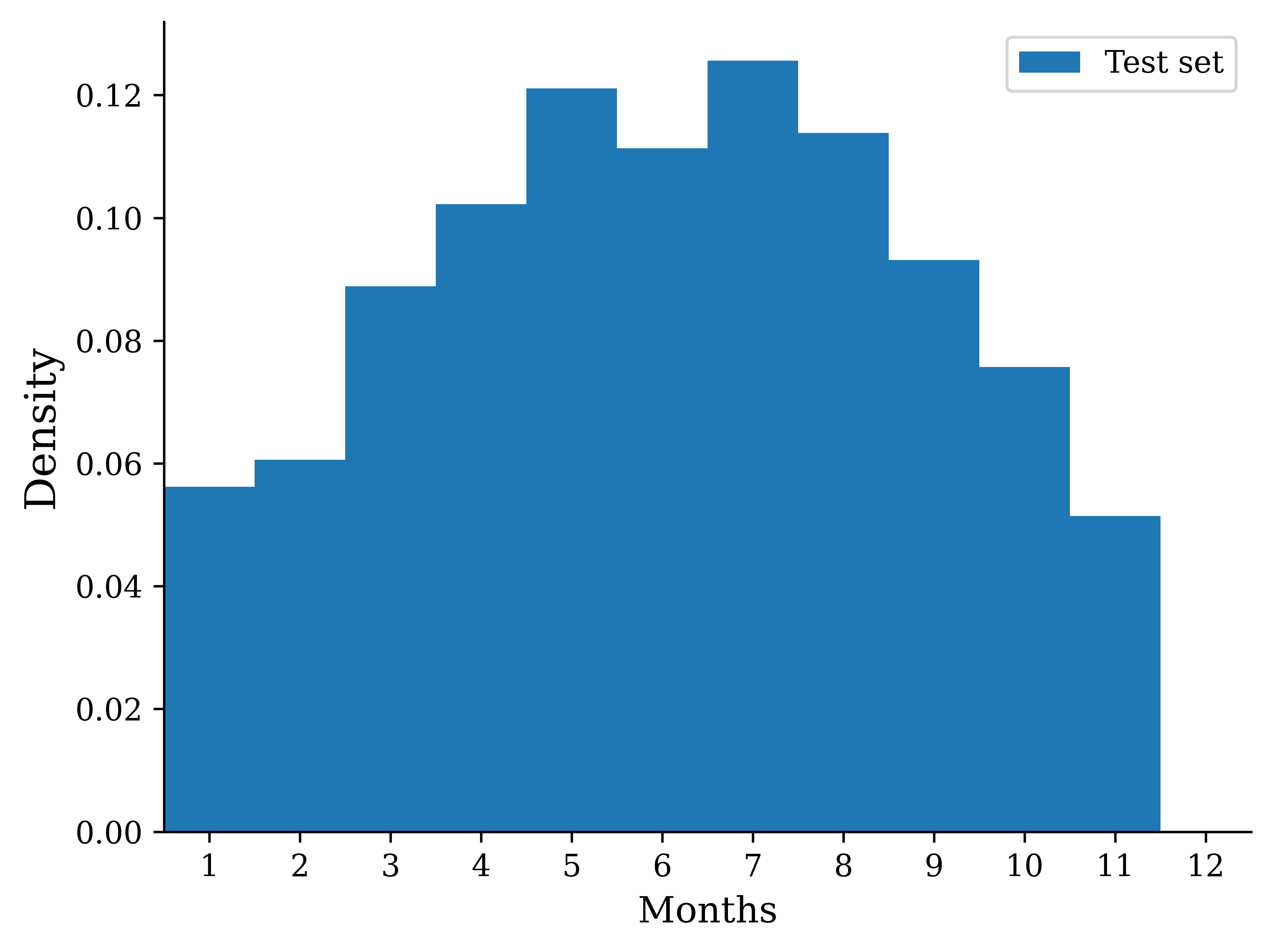}
    \label{fig:hist_months_test}
  \end{minipage}
  \vspace{-2\baselineskip}
\caption{Distribution of samples by months in the training, validation and test sets.}
\label{fig:dataset_distribution_month}
\end{figure}

  \vspace{-1\baselineskip}

\begin{figure}[h!]
\centering
\begin{minipage}[b]{0.33\textwidth}
    \includegraphics[width=1\textwidth]{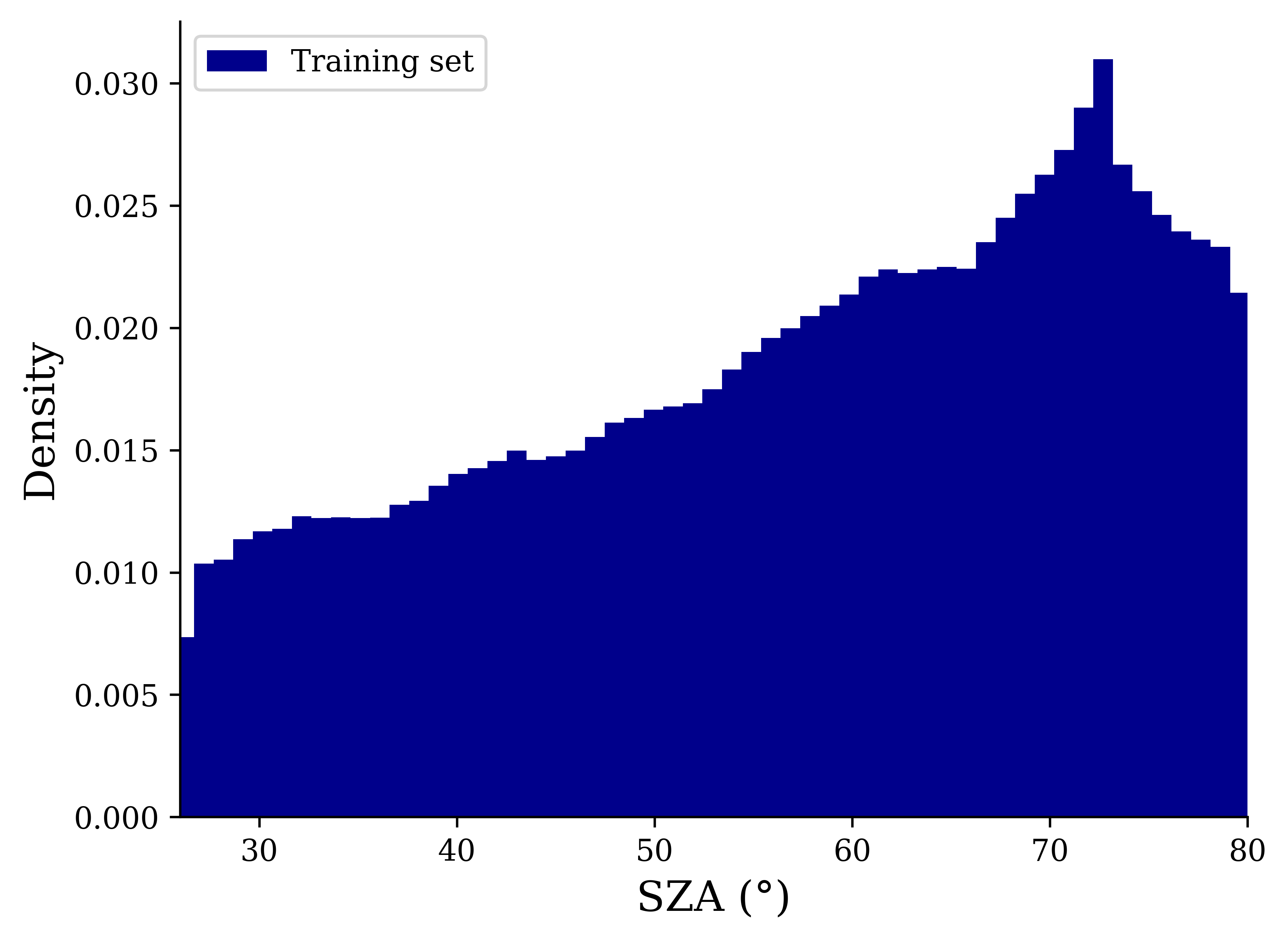}
    \label{fig:hist_sza_train}
  \end{minipage} 
  \begin{minipage}[b]{0.33\textwidth}
    \includegraphics[width=1\textwidth]{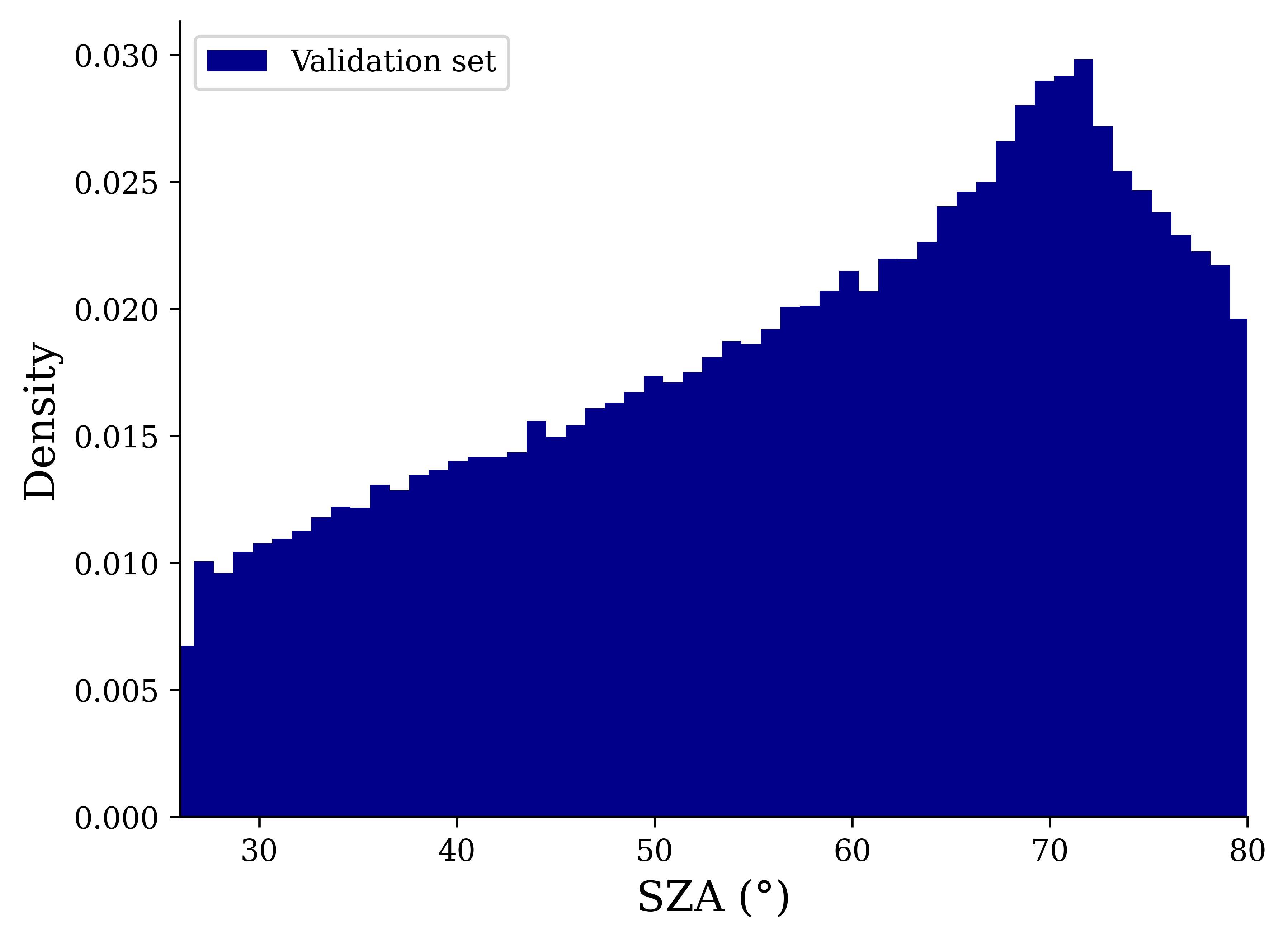}
    \label{fig:hist_sza_val}
  \end{minipage}
  \begin{minipage}[b]{0.33\textwidth}
    \includegraphics[width=1\textwidth]{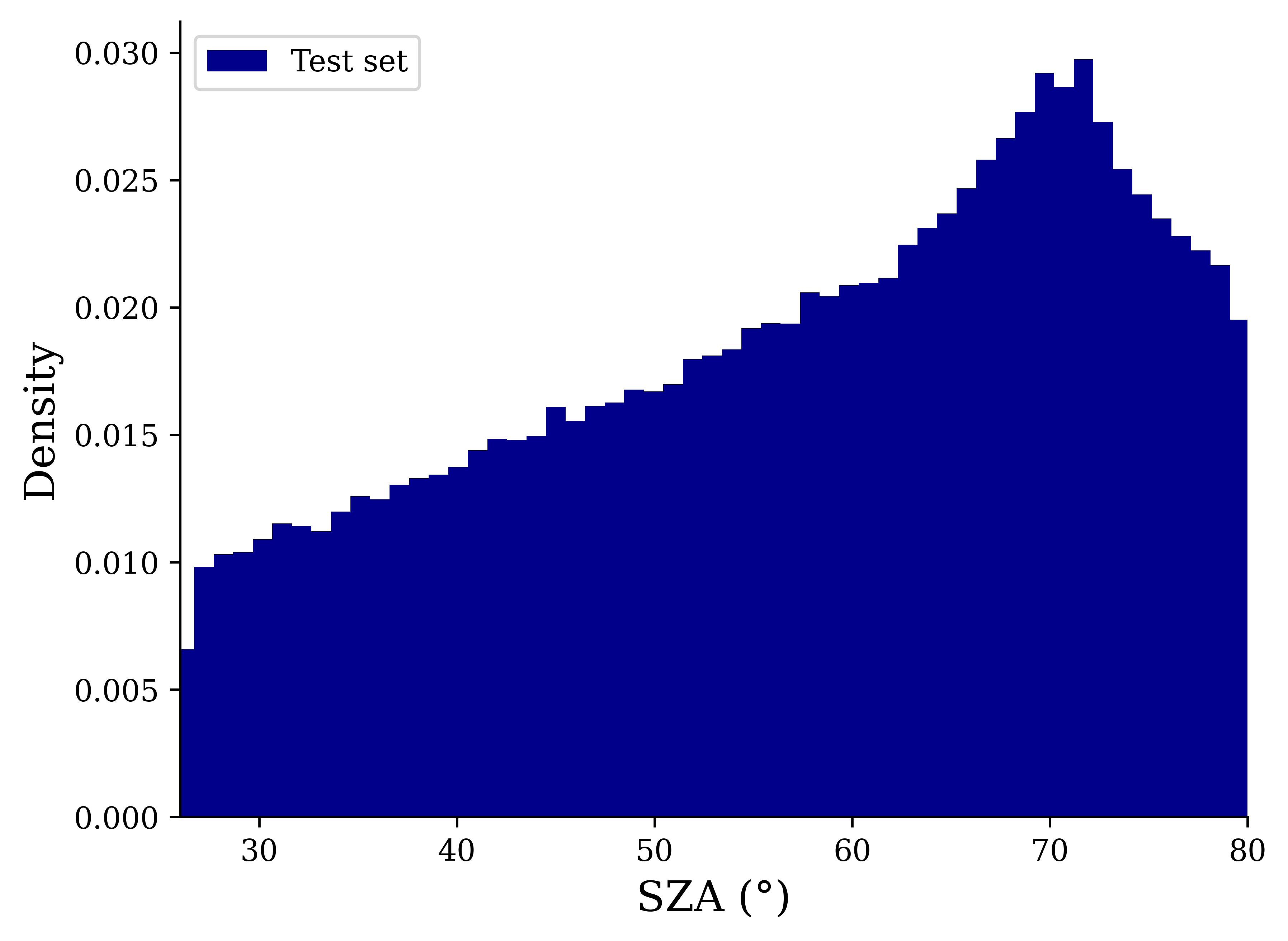}
    \label{fig:hist_sza_test}
  \end{minipage}
\vspace{-2\baselineskip}
\caption{Distribution of samples by Solar Zenith Angle in the training, validation and test sets.}
\label{fig:dataset_distribution_sza}
\end{figure}

\vspace{1\baselineskip}

\begin{figure}[h!]
\centering
\begin{minipage}[b]{0.33\textwidth}
    \includegraphics[width=1\textwidth]{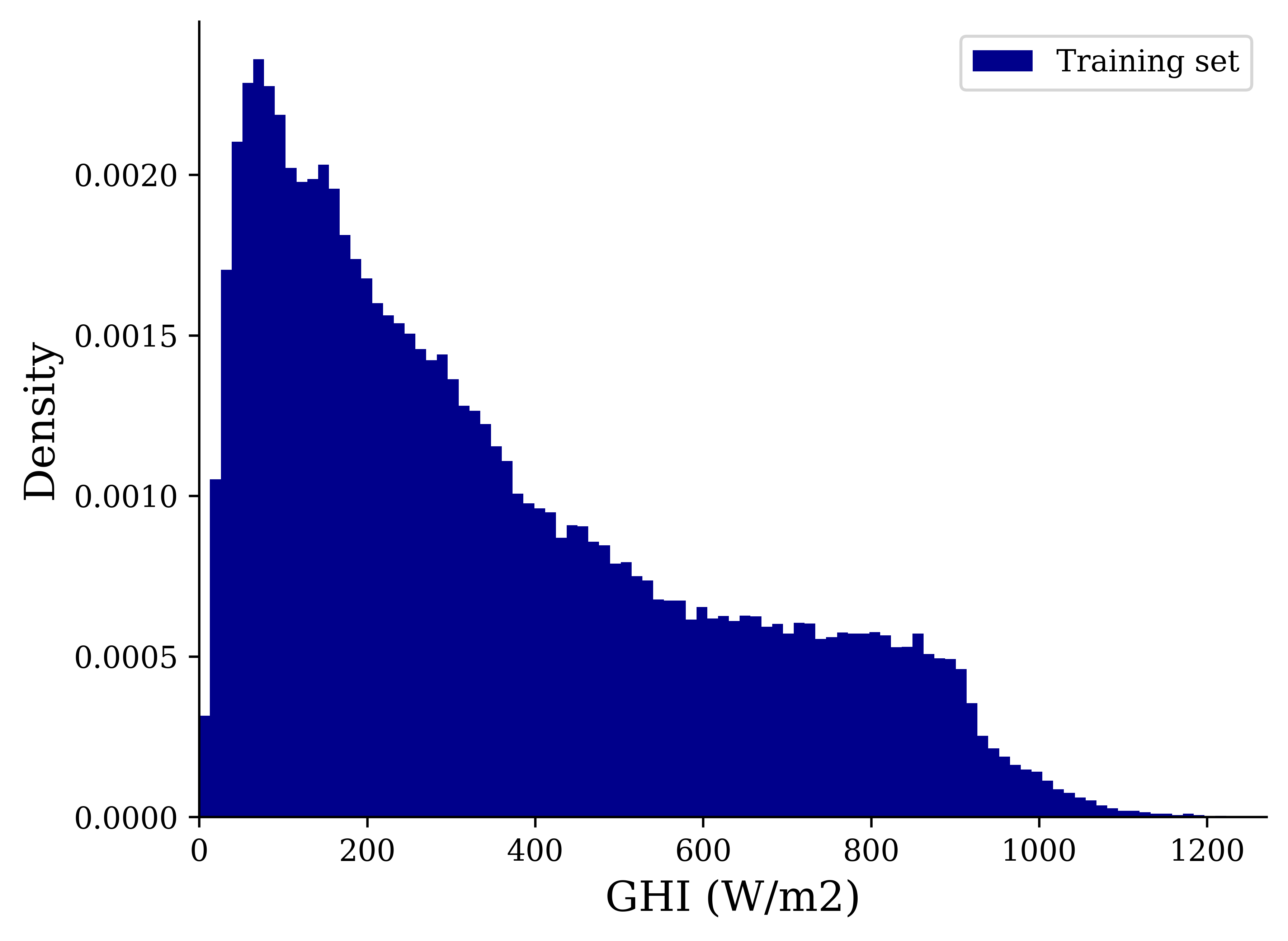}
    \label{fig:hist_bins_train}
  \end{minipage} 
  \begin{minipage}[b]{0.33\textwidth}
    \includegraphics[width=1\textwidth]{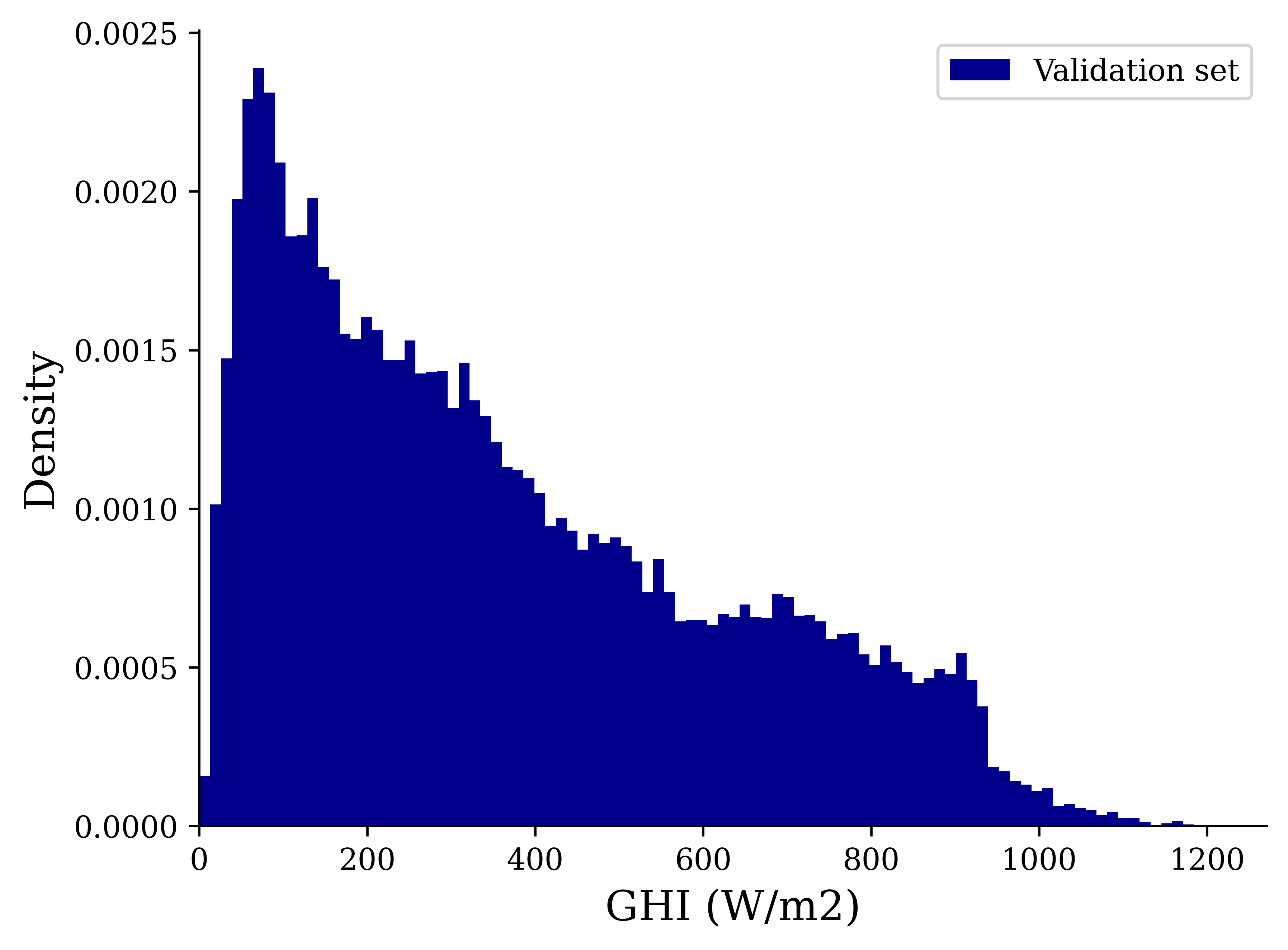}
    \label{fig:hist_bins_val}
  \end{minipage}
  \begin{minipage}[b]{0.33\textwidth}
    \includegraphics[width=1\textwidth]{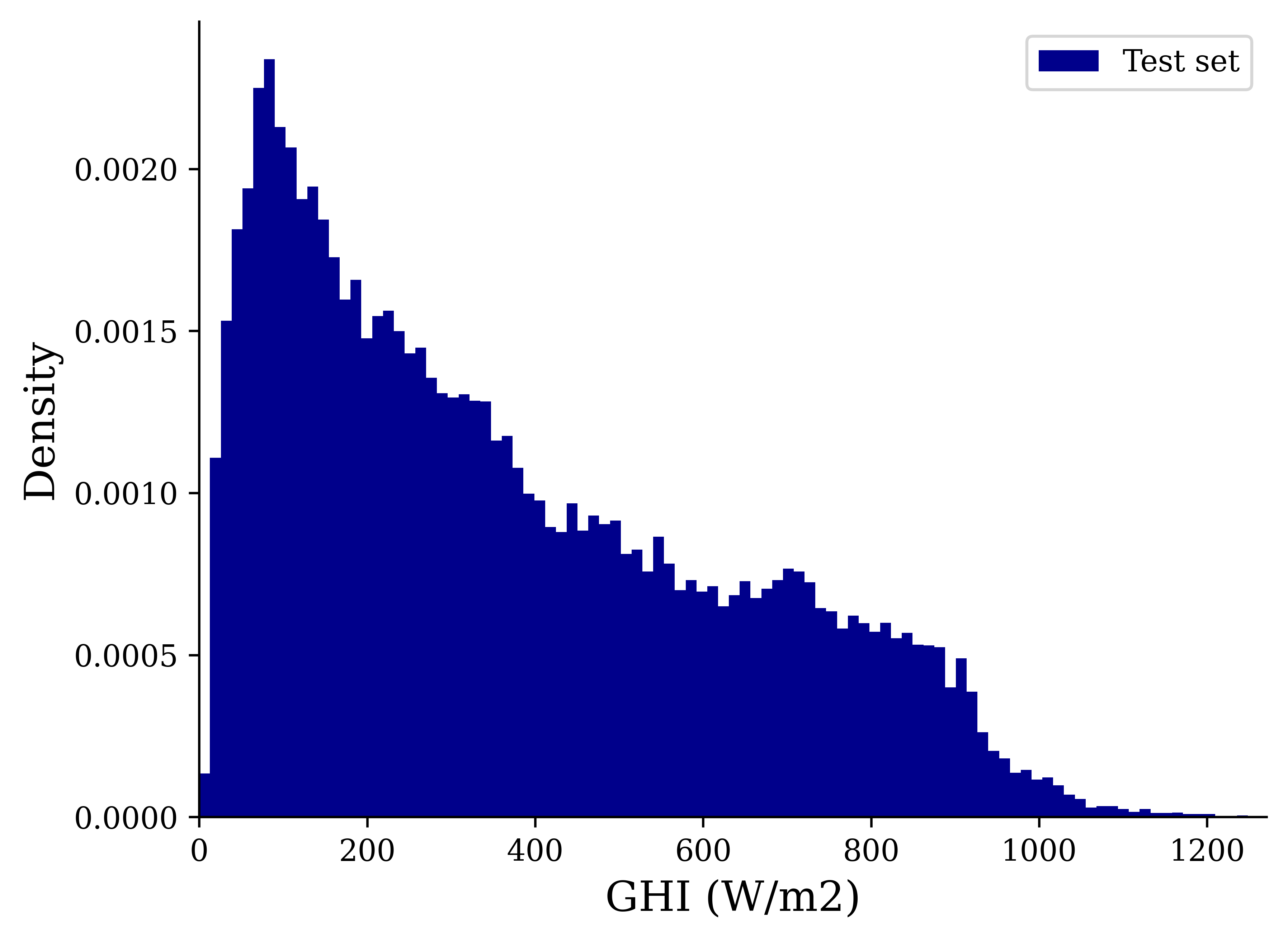}
    \label{fig:hist_bins_test}
  \end{minipage}
  \vspace{-2\baselineskip}
\caption{Distribution of samples by Global Horizontal Irradiance level in the training, validation and test sets.}
\label{fig:dataset_distribution_bins}
\end{figure}

\section{Forecasting setup}
\label{section:forecasting_setup}
\setcounter{figure}{0}
\setcounter{table}{0}

\begin{table*}[ht]
\begin{center}
\begin{tabular}{llcccc}
\hline
 & Model & Image Type & Resolution & Context & Forecasts  \\
\hline\hline
\noalign{\vskip 2mm}
Sky Image & CNN  & RGB & 128$\times$128 & 8 min (5 frames) & 2, 6, or 10 min  \\
 & ConvLSTM  & RGB & 128$\times$128 & 8 min (5 frames) &  2, 6, or 10 min\\
 & ECLIPSE  & RGB & 128$\times$128 & 8 min (5 frames) & 2, 4, 6, 8 and 10 min (+ Seg.) \\
\noalign{\vskip 2mm}
 Satellite Image & ECLIPSE  & Gray + CI + Irradiance & 128$\times$128 & 40 min (5 frames) & 10, 20, 30, 40 and 50 min (+ Seg.) \\
\noalign{\vskip 1mm}
\hline
\end{tabular}
\end{center}
\vspace{-1\baselineskip}
\caption{Features of the three models compared in this study.}
\label{tab:models}
\end{table*}

\newpage

\section{Auxiliary Data}
\label{section:auxiliary_data}
\setcounter{figure}{0}
\setcounter{table}{0}

\begin{table*}[ht!]
\begin{center}
\begin{tabular}{lcccccccc}
\hline
 && \multicolumn{3}{c}{Forecast Skill $\uparrow$ [\%]} && \multicolumn{3}{c}{TDI (Advance / Late)}\\

 ECLIPSE & $\mid$ & 10-min & 30-min & 50-min & $\mid$ & 10-min & 30-min & 50-min \\
\hline\hline
\vspace{0.5\baselineskip}
Absolute irradiance && -22.1\% & -2.6\% & 2.0\% && 18.2\% (8.5/9.5) & 18.5\% (8.8/9.7) & \textbf{19.4\%} (8.6/\textbf{10.8}) \\
Irradiance change && -0.6\% & -2.7\% & -3.8\% && \textbf{7.0\% (0.7/6.3)} & \textbf{16.8\%} (\textbf{2.8}/14.0) & 22.5\% (\textbf{4.5}/17.9) \\
Irradiance channel && 6.5\% & 6.2\% & 6.0\% && 10.5\% (2.5/7.9) & 17.8\% (4.4/13.4) & 20.1\% (7.5/12.7) \\
\begin{tabular}{@{}c@{}}Irradiance change \\ + Irradiance channel\end{tabular} && \textbf{9.4\%} &  \textbf{7.5\%} & \textbf{9.2\%} && 7.6\% (1.1/6.4) & 16.9\% (4.4/\textbf{12.5}) & 20.8\% (6.3/14.5) \\
\hline
\end{tabular}
\end{center}
\vspace{-1.2\baselineskip}
\caption{Predictions based on past irradiance measurements (with an additional irradiance channel or through predicting irradiance change) significantly outperform forecasts from satellite images only (absolute irradiance). Overall, combining irradiance change prediction with an additional irradiance input channel provides the best FS on all forecast horizons. All sequences (predictions and ground truth) are normalised prior to determining the temporal distortion metrics.}
\label{tab:ablation_study_aux_data}
\end{table*}

\section{Cloud Index Map}
\label{section:cloud_index_map}
\setcounter{figure}{0}
\setcounter{table}{0}
\setcounter{equation}{0}

A traditional preprocessing step in satellite imagery for solar applications is to remove background elements (water, lands, etc.) to determine a cloud map from raw images. The resulting cloud map can be ultimately turned into an irradiance map given the physical properties of visible clouds and a clear-sky model~\cite{blancHelioClimProjectSurface2011}.

\vspace{0.5\baselineskip}

A common way to define a cloud map is to use the cloud index (CI). Ranging from 0 (no cloud) to 1 (thick cloud), this index can be computed given the statistics of the pixel values (Equation~\ref{equ:cloud_index}): $p(i, j, t)$ being the value of a given pixel ($i,j$) at time $t$; $p_{min}(i, j, t-N:t)$ the minimum pixel value of the same pixel ($i,j$) at the same time $t$ of the day over the last $N$ last days ($N=10$ here) and $p_{max}(t)$ the maximum pixel value in the image at time $t$.

\begin{equation}
   \text{cloud index} (i, j, t) = \frac{p(i, j, t)- p_{min}(i, j, t-N:t)}{p_{max}(t) - p_{min}(i, j, t-N:t)}
   \label{equ:cloud_index}
\end{equation}

\vspace{0.5\baselineskip}

In practice, $p_{min}(i, j, t-N:t)$ corresponds to the solar radiation reflected by the ground (albedo) when it is not occluded by clouds. On the contrary, $p_{max}(t)$ correlates with the highest solar radiation reflected by visible clouds at a given time. These two references define the range of values that a pixel value can take (from $p_{min}(i, j, t-N:t)$ to $p_{max}(t)$) given the local state of the cloud cover (from no cloud to thick clouds). The corresponding segmentation classifies each pixel into 5 classes given its cloud index: 0-20\%, 20-40\%, 40-60\%, 60-80\% and 80-100\%.

\vspace{0.5\baselineskip}

\begin{figure*}[h!]
\centering
\begin{minipage}[b]{0.245\textwidth}
    \includegraphics[width=1\textwidth]{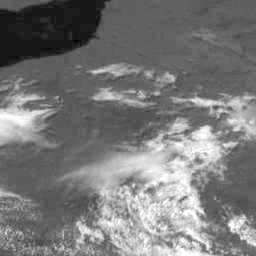}
    \label{fig:cloud_index_raw}
  \end{minipage}
  \begin{minipage}[b]{0.245\textwidth}
    \includegraphics[width=1\textwidth]{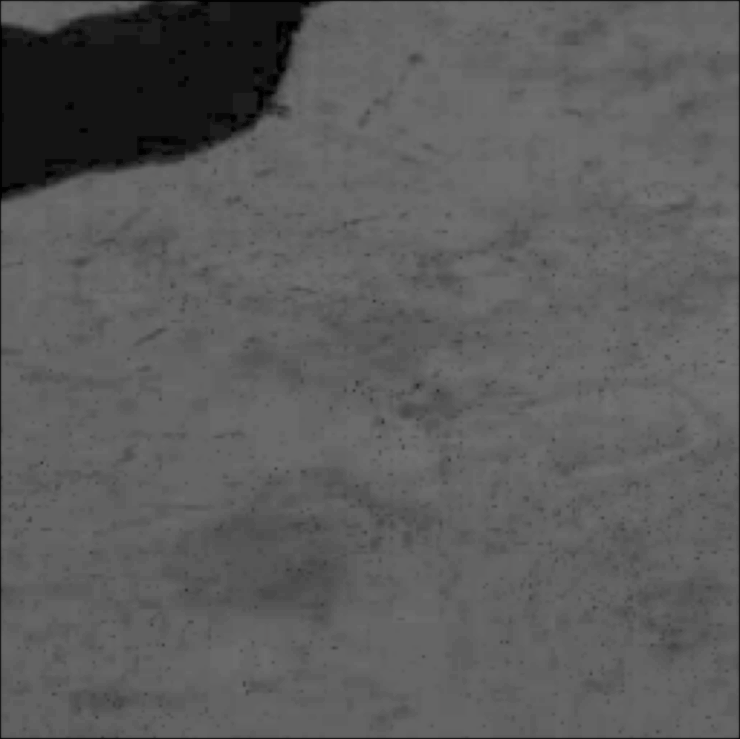}
    \label{fig:cloud_index_background}
  \end{minipage}
   \begin{minipage}[b]{0.245\textwidth}
    \includegraphics[width=1\textwidth]{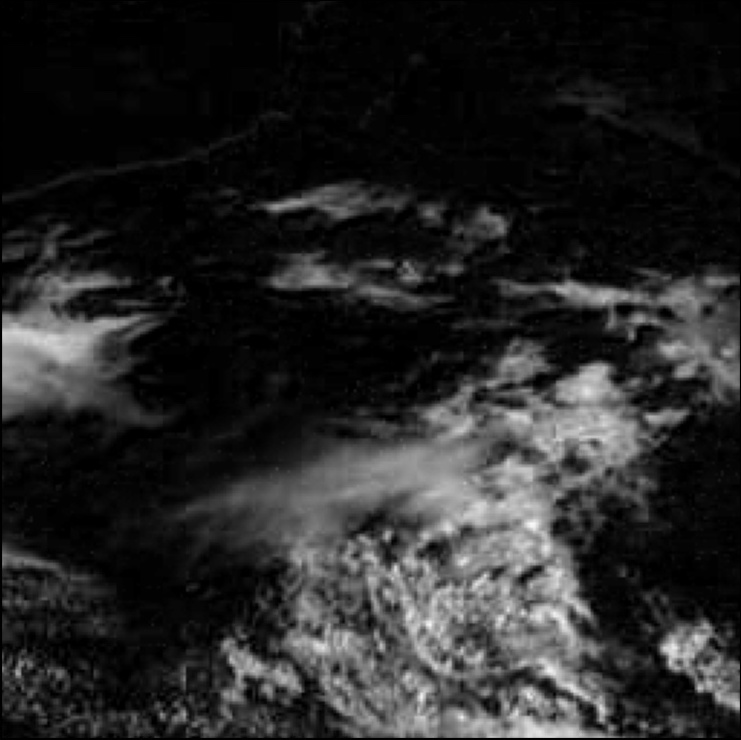}
    \label{fig:cloud_index}
  \end{minipage} 
  \begin{minipage}[b]{0.245\textwidth}
    \includegraphics[width=1\textwidth]{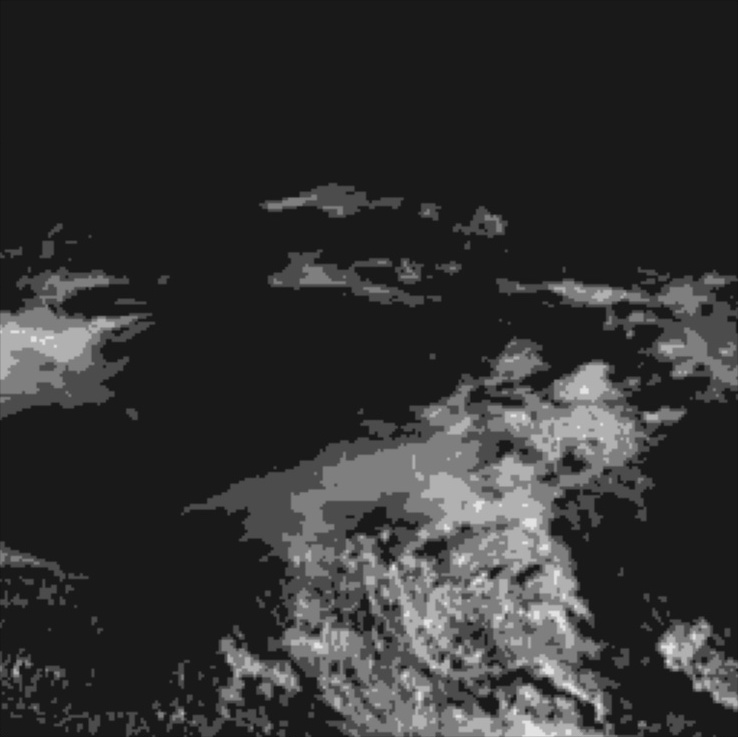}
    \label{fig:cloud_index_segmentation}
\end{minipage}

\vspace{-1.2\baselineskip}
\caption{From left to right: 1. Raw image, 2. Background ($p_{min}(t-N:t)$), 3. Cloud index map, 4. Segmentation of the cloud index map. Notice that the difference between sea and land is not visible in panels 3 and 4.}
\label{fig:cloud_index_maps}
\end{figure*}

Compared to raw images, the cloud index maps are not affected by the diurnal change of lighting conditions. This provides less information on the level of irradiance on the ground but focuses instead on the local transmittance of the cloud cover (Figure~\ref{fig:cloud_index_maps}).

\newpage

\section{Results - Sky Images \& Satellite Images}
\label{section:results}
\setcounter{figure}{0}
\setcounter{table}{0}

\begin{table*}[ht]
\begin{center}
\begin{tabular}{llcccccccc}
\hline
 &&& \multicolumn{3}{c}{RMSE $\downarrow$ [W/$\text{m}^2$] (Forecast Skill $\uparrow$ [\%])} & & \multicolumn{3}{c}{95\% Quantile $\downarrow$ [W/$\text{m}^2$]}\\
\noalign{\vskip 0.7mm}
 Model & Transformation & $\mid$ & 2-min & 6-min & 10-min & $\mid$ & 2-min & 6-min & 10-min \\
 \noalign{\vskip 0.5mm}
\hline\hline
\noalign{\vskip 2mm}
Smart Pers. &&& 93.3 (0\%) & 129.7 (0\%) & 146.2 (0\%) & & 201.1 & 304.8 & 356.9 \\
\noalign{\vskip 2mm}
\hline
\noalign{\vskip 2mm}
CNN & Raw && 83.0 (11.1\%) & 104.8 (19.2\%) & 115.1 (21.2\%) && 182.2 & 241.9 & 268.3 \\
 & Raw + Rotations && 76.6 (17.9\%) & 97.1 (25.1\%) & 108.7 (25.6\%) && 168.5 & 222.4 & 251.9 \\
\\
 & CSA && 82.3 (11.8\%) & 102.4 (21.1\%) & 115.4 (21.1\%) && 183.0 & 233.5 & 264.3 \\
 & CSA + Rotations && \textbf{71.4 (23.5\%)} & 94.0 (27.5\%) & 109.0 (25.4\%) && \textbf{152.7} & 215.7 & 252.4 \\
\\
 & Sun-Centred && 82.4 (11.7\%) & 101.9 (21.4\%) & 112.7 (22.9\%) && 181.3 & 236.1 & 260.7 \\
 & Sun-Centred + Rotations && 75.5 (19.1\%) & 97.2 (25.1\%) & 109.5 (25.1\%) && 163.0 & 220.3 & 250.9 \\
\\
 & SPIN && 80.7 (13.5\%) & 101.9 (21.5\%) & 112.2 (23.3\%) && 180.2 & 236.3 & 257.4 \\
 & SPIN + Translations && 72.5 (22.3\%) & \textbf{93.3 (28.0\%)} & \textbf{107.3 (26.6\%)} && 154.8 & \textbf{211.7} & \textbf{247.0} \\

\noalign{\vskip 2mm}
\hline
\noalign{\vskip 2mm}

ConvLSTM & Raw && 81.2 (13.0\%) & 99.4 (23.3\%) & 111.4 (23.8\%) && 177.6 & 227.0 & 256.1 \\
 & Raw + Rotations && 78.0 (16.5\%) & 93.0 (28.3\%) & 107.3 (26.6\%) && 171.1 & 210.6 & 247.2 \\
\\
 & CSA && 76.0 (18.6\%) & 100.0 (22.9\%) & 113.0 (22.7\%) && 168.4 & 228.4 & 263.2 \\
 & CSA + Rotations && \textbf{69.5 (25.5\%)} & 92.6 (28.6\%) & 107.2 (26.7\%) && \textbf{153.3} & 210.6 & 246.7 \\
\\
 & Sun-Centred && 77.9 (16.5\%) & 98.6 (25.0\%) & 109.4 (25.2\%) && 170.3 & 225.8 & 254.7 \\
 & Sun-Centred + Rotations && 72.7 (22.1\%) & \textbf{91.0 (29.8\%)} & \textbf{105.8 (27.6\%)} && 158.1 & \textbf{206.8} & \textbf{242.9} \\
\\
 & SPIN && 74.3 (20.3\%) & 96.3 (25.7\%) & 109.5 (25.1\%) && 162.9 & 222.2 & 254.7 \\
 & SPIN + Translations && 71.3 (23.6\%) & 91.5 (29.4\%) & 107.7 (26.4\%) && \textbf{153.9} & 209.3 & 248.6 \\

\noalign{\vskip 2mm}
\hline
\noalign{\vskip 2mm}

ECLIPSE & Raw && 85.4 (8.4\%) & 98.3 (24.2\%) & 110.4 (24.5\%) && 184.9 & 218.2 & 244.8 \\
 & Raw + Rotations && 78.6 (15.8\%) & 93.6 (27.8\%) & 105.5 (27.8\%) && 168.6 & 205.9 & 238.6 \\
\\
 & CSA && 80.4 (13.8\%) & 97.0 (25.5\%) & 109.8 (25.1\%) && 169.9 & 217.4 & 244.1 \\
 & CSA + Rotations && 77.1 (17.3\%) & 93.5 (27.9\%) & 107.3 (26.6\%) && 161.2 & 205.9 & 240.0 \\
\\
 & Sun-Centred && 81.0 (13.2\%) & 94.1 (27.4\%) & 106.6 (27.1\%) && 174.4 & 206.8 & 235.1 \\
 & Sun-Centred + Rotations && 74.9 (19.8\%) & 89.2 (31.2\%) & 102.7 (29.8\%) && 157.8 & 196.1 & 227.4 \\
\\
 & SPIN && 74.3 (20.4\%) & 89.3 (31.1\%) & 102.9 (29.6\%) && 156.6 & 196.6 & 230.3 \\
 & SPIN + Translations && \textbf{71.8 (23.1\%)} & \textbf{87.2 (32.8\%)} & \textbf{101.0 (30.9\%)} && \textbf{149.1} & \textbf{192.2} & \textbf{224.0} \\

\noalign{\vskip 2mm}
\hline
\end{tabular}
\end{center}
\vspace{-1\baselineskip}
\caption{Comparison of the different image transformations based on the model performance on the 2, 6 and 10-min ahead irradiance predictions from sky images. Each score is reported as the average over two trainings with different random initialisations. We observe that polar transformations and data augmentation significantly improve predictions.}
\label{tab:results_table}
\end{table*}

\begin{table*}[ht]
\begin{center}
\begin{tabular}{lcccccccc}
\hline
 & & \multicolumn{3}{c}{RMSE $\downarrow$ [W/$\text{m}^2$] (Forecast Skill $\uparrow$ [\%])} & & \multicolumn{3}{c}{95\% Quantile $\downarrow$ [W/$\text{m}^2$]}\\

 Forecast Horizon & $\mid$ & 10-min & 30-min & 50-min & $\mid$ & 10-min & 30-min & 50-min \\
\hline\hline
\noalign{\vskip 2mm}
\multicolumn{2}{l}{Smart Pers.} & 123.5 (0\%) & 146.7 (0\%) & 162.2 (0\%) && 295.5 & 350.2 & 382.5 \\
\noalign{\vskip 2mm}
Raw && 113.1 (8.5\%) & 137.2 (6.5\%) & 150.5 (6.3\%) && 260.7 & 305.0 & 326.7 \\
Raw + Rotations && 108.6 (12.1\%) & 128.8 (12.2\%) & 138.7 (13.7\%) && 253.1 & 290.1 & 307.4 \\
\noalign{\vskip 2mm}
Close-up && 110.8 (10.3\%) & 131.2 (10.6\%) & 141.8 (11.8\%) && 254.0 & 297.8 & 311.1 \\
Close-up + Rotations && 110.0 (10.9\%) & 130.4 (11.1\%) & 131.8 (18.0\%) && 255.6 & 290.3 & \textbf{290.7} \\
\noalign{\vskip 2mm}
SPIN && 108.4 (12.2\%) & 128.4 (12.5\%) & 138.0 (14.1\%) && 249.5 & 292.1 & 302.6 \\
SPIN + Translations && 108.5 (12.1\%) & 129.6 (13.9\%) & 134.5 (17.1\%) && 254.1 & 297.1 & 302.7 \\
\noalign{\vskip 2mm}
SPIN Close-up && \textbf{106.5 (13.8\%)} & 125.8 (14.3\%) & 136.9 (14.8\%) && 249.4 & \textbf{283.8} & 304.5 \\
SPIN Close-up + Translations && 106.7 (13.6\%) & \textbf{125.1 (14.4\%)} & \textbf{129.0 (18.0\%)} && \textbf{247.0} & 288.6 & 291.9 \\
\noalign{\vskip 1mm}
\hline
\end{tabular}
\end{center}
\vspace{-1\baselineskip}
\caption{Comparison of performance of the different satellite image transformations based on ECLIPSE 10, 30 and 50-min ahead irradiance predictions averaged over a 5-min period.}
\label{tab:satellite_results_eclipse_long}
\end{table*}

\newpage

\section{Examples of a predicted GHI profile (2-min resolution)}
\label{section:prediction_curves}
\setcounter{figure}{0}
\setcounter{table}{0}

\begin{figure*}[ht]
\centering    
\includegraphics[width=0.9\textwidth]{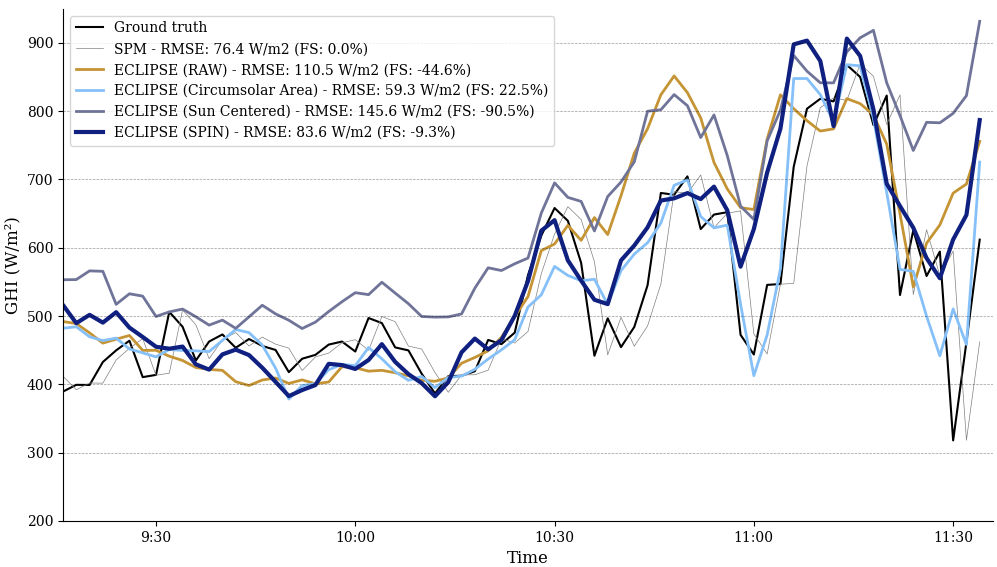}
\vspace{-0.5\baselineskip}
\caption{2-min ahead prediction curves of ECLIPSE for different image transformations of sky images (27/08/2019). Transformations magnifying the sun region (SPIN and CSA) suffer less from a consistent bias in some weather conditions. Despite anticipation skills, this underestimation of the irradiance level significantly increases the average error of the models trained on raw and sun-centred images.}
\label{fig:prediction_curves_1lf}
\end{figure*}

\begin{figure*}[ht]
\centering    
\includegraphics[width=0.9\textwidth]{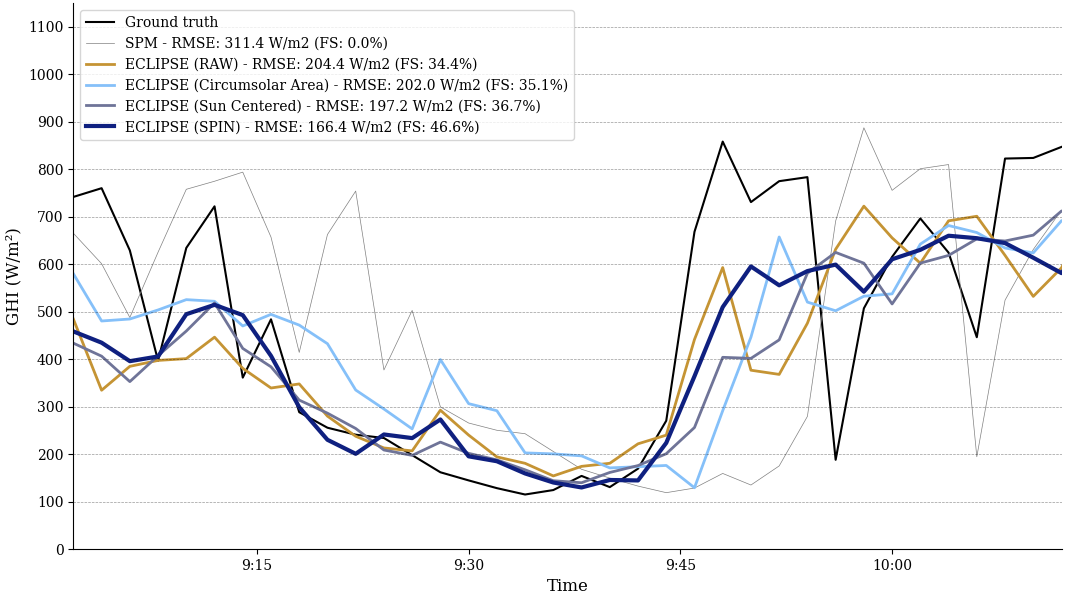}
\vspace{-0.5\baselineskip}
\caption{10-min ahead prediction curves of ECLIPSE for different image transformations of sky images (03/04/2019). The limited field of view induced by the close-up on the sun in the CSA transform decreases the anticipation skills of the model.}
\label{fig:prediction_curves_5lf}
\end{figure*}

\clearpage
\newpage

\section{Principal Component Analysis on the Spatiotemporal Representation of the Input Sequence of Past Sky Images}
\label{section:pca}
\setcounter{figure}{0}
\setcounter{table}{0}

\begin{figure}[ht!]
     \centering
\begin{subfigure}[b]{1\textwidth}
\centering
\begin{minipage}[b]{0.19\textwidth}
    \includegraphics[width=1\textwidth]{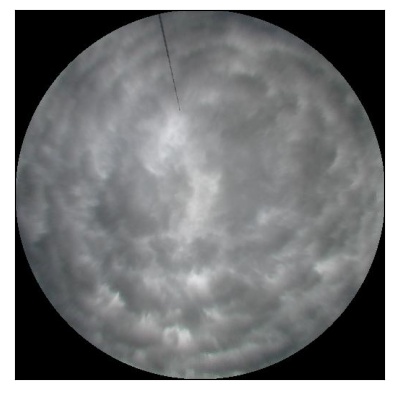}
  \end{minipage} 
\begin{minipage}[b]{0.19\textwidth}
    \includegraphics[width=1\textwidth]{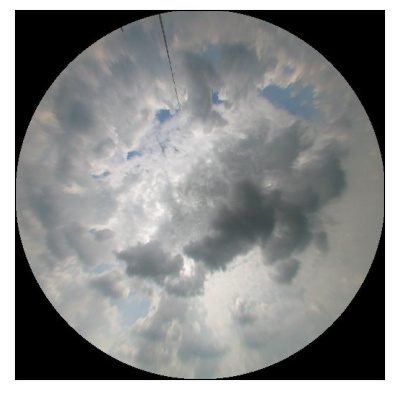}
  \end{minipage}
  \begin{minipage}[b]{0.19\textwidth}
    \includegraphics[width=1\textwidth]{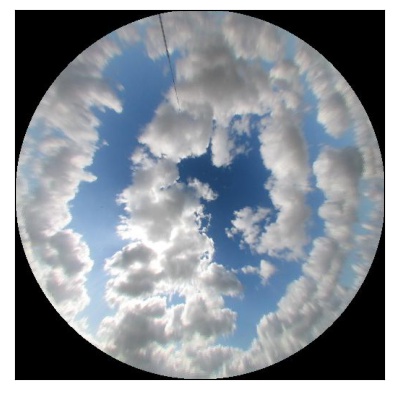}
  \end{minipage}
   \begin{minipage}[b]{0.19\textwidth}
    \includegraphics[width=1\textwidth]{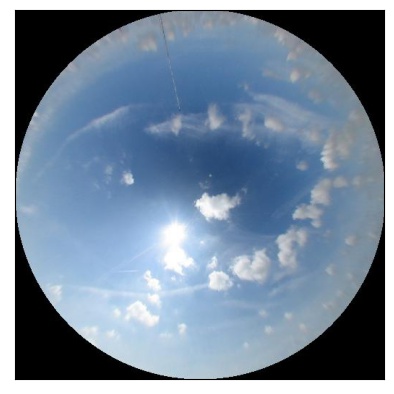}
  \end{minipage} 
  \begin{minipage}[b]{0.19\textwidth}
    \includegraphics[width=1\textwidth]{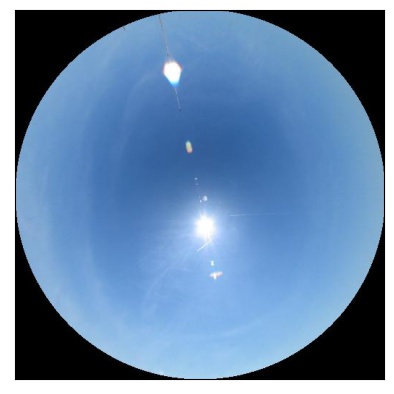}
\end{minipage}
\caption{PC1: extent of the cloud coverage.}
\label{fig:pc1_stereo}
\end{subfigure}
\hfill
\vspace{1\baselineskip}
\begin{subfigure}[b]{1\textwidth}
\centering
\begin{minipage}[b]{0.19\textwidth}
    \includegraphics[width=1\textwidth]{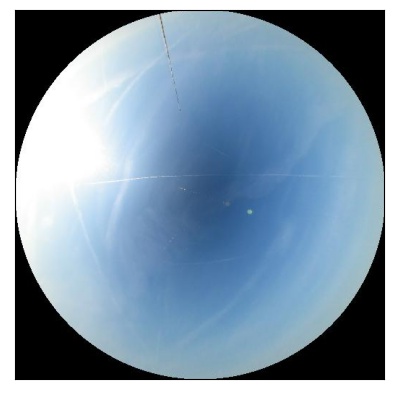}
  \end{minipage} 
\begin{minipage}[b]{0.19\textwidth}
    \includegraphics[width=1\textwidth]{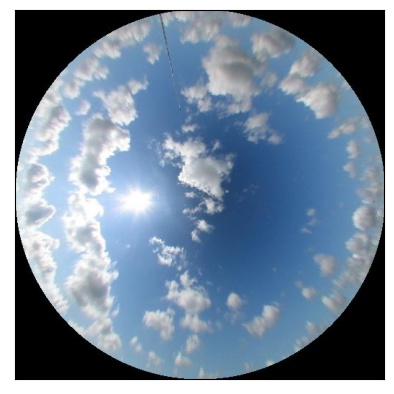}
  \end{minipage}
  \begin{minipage}[b]{0.19\textwidth}
    \includegraphics[width=1\textwidth]{Figures/pca_nb7770_img_201904171036.jpg}
  \end{minipage}
   \begin{minipage}[b]{0.19\textwidth}
    \includegraphics[width=1\textwidth]{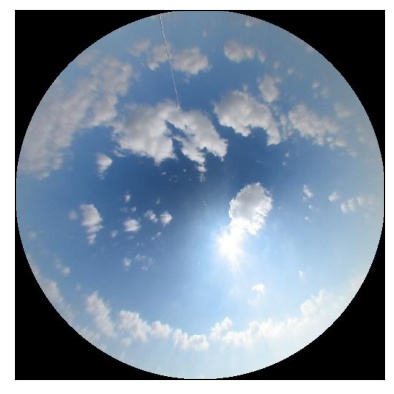}
  \end{minipage} 
  \begin{minipage}[b]{0.19\textwidth}
    \includegraphics[width=1\textwidth]{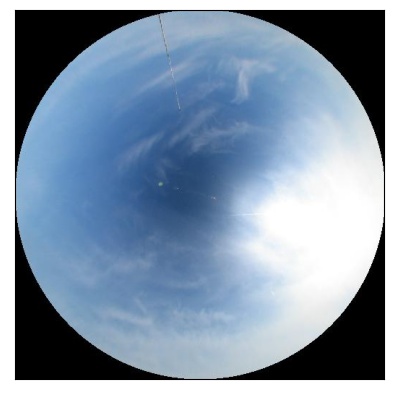}
\end{minipage}
\caption{PC2: horizontal position of the sun in the sky.}
\label{fig:pc2_stereo}
\end{subfigure}
\hfill
\vspace{1\baselineskip}
\begin{subfigure}[b]{1\textwidth}
\centering
\begin{minipage}[b]{0.19\textwidth}
    \includegraphics[width=1\textwidth]{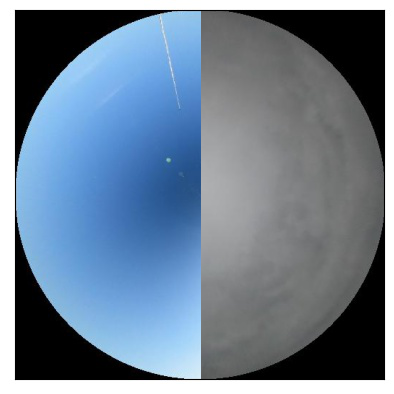}
  \end{minipage} 
\begin{minipage}[b]{0.19\textwidth}
    \includegraphics[width=1\textwidth]{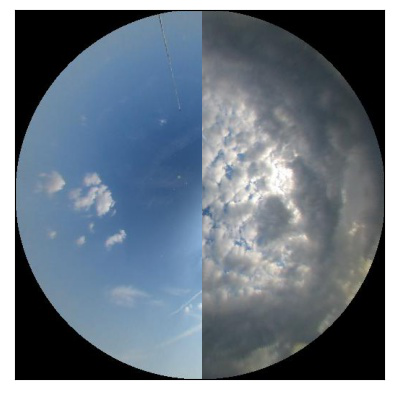}
  \end{minipage}
  \begin{minipage}[b]{0.19\textwidth}
    \includegraphics[width=1\textwidth]{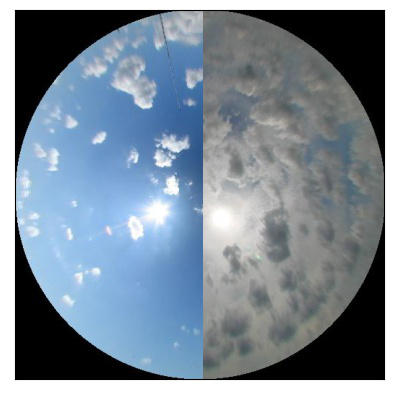}
  \end{minipage}
   \begin{minipage}[b]{0.19\textwidth}
    \includegraphics[width=1\textwidth]{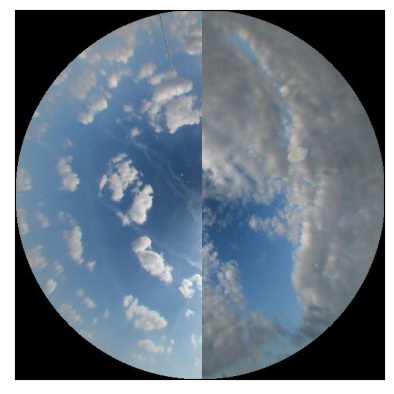}
  \end{minipage} 
  \begin{minipage}[b]{0.19\textwidth}
    \includegraphics[width=1\textwidth]{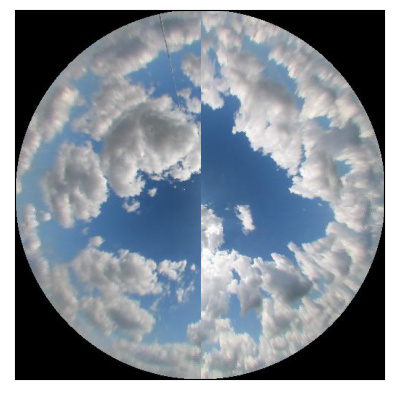}
\end{minipage}
\caption{PC3: spatial variability of the cloud cover: from fully cloudy or fully sunny to partly cloudy. Each image illustrates two neighbouring samples from the distribution.}
\label{fig:pc3_stereo}
\end{subfigure}
\hfill
\vspace{1\baselineskip}
\begin{subfigure}[b]{1\textwidth}
\centering
\begin{minipage}[b]{0.19\textwidth}
    \includegraphics[width=1\textwidth]{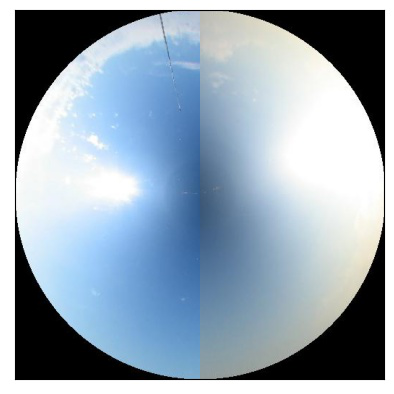}
  \end{minipage} 
\begin{minipage}[b]{0.19\textwidth}
    \includegraphics[width=1\textwidth]{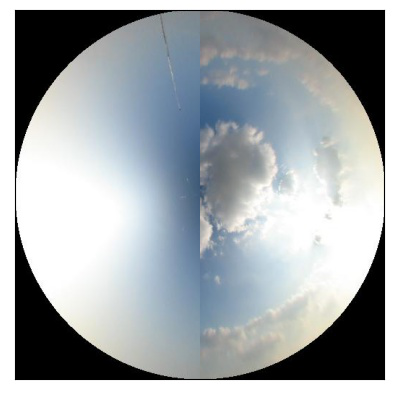}
  \end{minipage}
  \begin{minipage}[b]{0.19\textwidth}
    \includegraphics[width=1\textwidth]{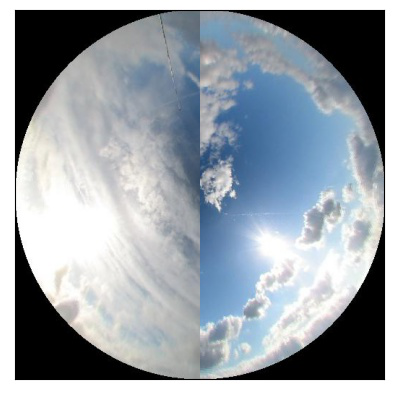}
  \end{minipage}
   \begin{minipage}[b]{0.19\textwidth}
    \includegraphics[width=1\textwidth]{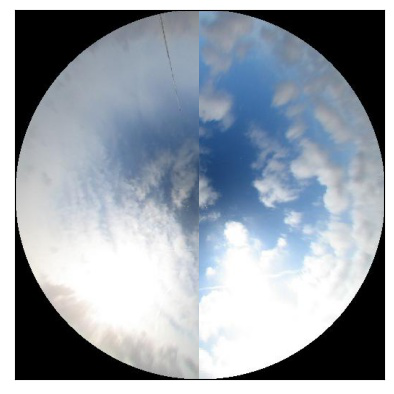}
  \end{minipage} 
  \begin{minipage}[b]{0.19\textwidth}
    \includegraphics[width=1\textwidth]{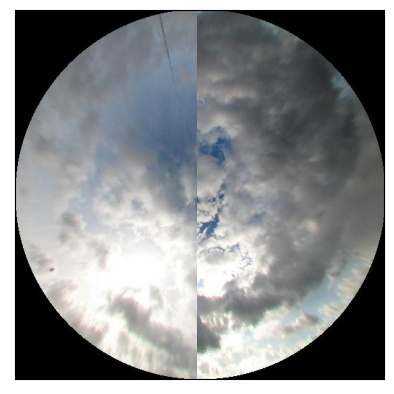}
\end{minipage}
\caption{PC4: vertical position of the sun in the sky. Each image illustrates two neighbouring samples from the distribution.}
\label{fig:pc4_stereo}
\end{subfigure}
\vspace{1\baselineskip}
\caption{Four principal components of the spatiotemporal representation encoded by the model for raw sky images. The variability of each component is illustrated with images drawn from the distribution, from low to high values. Taken from~\cite{palettaECLIPSEEnvisioningCloud2021}.}
\label{fig:pca_components_examples_stereo}
\end{figure}

\end{document}